%% file: glob3r.tex
\documentclass{article}
\PassOptionsToPackage{numbers,sort&compress}{natbib}

\usepackage{glob3r,times}

\usepackage{xspace}
\usepackage{amsmath}
\usepackage{enumitem}
\usepackage[utf8]{inputenc} 
\usepackage[T1]{fontenc}    
\usepackage{hyperref}       
\usepackage{url}            
\usepackage{booktabs}       
\usepackage{amsfonts}       
\usepackage{nicefrac}       
\usepackage{microtype}      
\usepackage{xcolor}         
\usepackage{graphicx}
\usepackage{multirow} 
\setcitestyle{numbers,square,comma}
\title{Glob3R: Global Structure-from-Motion with 3D Foundation Models}

\author{
\quad\quad \quad Junyuan Deng\textsuperscript{1,2}\thanks{Equal contribution.}\quad
Heng Li\textsuperscript{1}\footnotemark[1]\quad
Kejie Qiu\textsuperscript{2}\footnotemark[1]\quad
Lingteng Qiu\textsuperscript{2}\quad
Rui Peng\textsuperscript{2}\\
\quad\quad \quad\textbf{Weichao Shen\textsuperscript{2}\quad
Weihao Yuan\textsuperscript{3}\quad
Siyu Zhu\textsuperscript{4}\quad
Zilong Dong\textsuperscript{2}\quad
Ping Tan}\textsuperscript{1}\thanks{Corresponding authors.\hangindent=1.8em\hangafter=1\\E-mail: {pingtan@ust.hk}}
\\
\vspace{-0.4cm}
\and
\textsuperscript{1}{The Hong Kong University of Science and Technology}\quad
\textsuperscript{2}{Tongyi Lab, Alibaba Group}\\
\textsuperscript{3}{Nanjing University}\quad
\textsuperscript{4}{Fudan University}
\vspace{3pt}\\
}

\iclrfinalcopy 

\begin{document}

\maketitle

\input{sections/00_abstract}

\input{sections/01_intro}

\input{sections/02_related_works}

\input{sections/03_method}

\input{sections/04_experiment}

\input{sections/05_conclusion}

{
    \small
    \bibliographystyle{plainnat}
    \bibliography{main}
}

\newpage
\input{sections/XX_supp}

\end{document}

%% file: sections/00_abstract.tex
\begin{abstract}
\vspace{2mm}
Recent 3D geometric foundation models, such as VGGT, provide robust feed-forward 3D reconstruction by directly predicting camera poses and 3D scene points from input images. However, their results remain inaccurate, and scaling them to long sequences or large unordered image sets typically requires chunk-wise processing, which can introduce drift and inconsistency. We present Glob3R, a global SfM-style reconstruction built on 3D foundation models. Our key idea is to explicitly optimize feed-forward geometric predictions. To this end, we augment a frozen Pi3X backbone with a lightweight dense matching head that predicts image warps between selected reference frames and neighboring views. These dense warps are converted into sparse but reliable multi-view feature tracks, which provide correspondence constraints for global optimization. We further introduce a keyframe-based sliding-window association strategy that propagates tracks and relative poses across overlapping windows, enabling scalable reconstruction. Finally, we perform global motion averaging and bundle adjustment to refine camera poses, reduce scale inconsistencies, and recover dense scene geometry. Extensive experiments on indoor, outdoor, large-scale driving, and unordered SfM benchmarks demonstrate that Glob3R achieves robust and accurate reconstruction. It consistently improves over feed-forward foundation-model baselines and recent scalable reconstruction methods, while being more robust than classical SfM pipelines. The refined poses also lead to higher-quality neural rendering, validating the benefit of combining foundation-model priors with global geometric optimization. Project page:~\url{https://junyuandeng.github.io/Glob3r/}
\end{abstract}

%% file: sections/01_intro.tex
\vspace{3mm}

\section{Introduction}
\vspace{1mm}
Reconstructing 3D scenes from image collections remains a fundamental challenge in computer vision, serving as a cornerstone for applications such as augmented reality (AR), robotics, autonomous navigation, and neural rendering. Recently, learning-based 3D geometric foundation models, including DUSt3R~\cite{wang2024dust3r}, VGGT~\cite{wang2025vggt}, Pi3X~\cite{wang2025pi3}, and other recent methods~\cite{keetha2025mapanything, shen2025fastvggt, deng2025reloc, gao2025more, deng2025boost, yan2025360recon}, have emerged as a new reconstruction paradigm. Departing from traditional pipelines, these models directly predict camera poses and dense per-pixel geometry, namely depth or scene coordinates, from arbitrary image sets in a feed-forward manner. This capability offers a robust and highly efficient initialization for 3D reconstruction from both ordered image sequences and unordered image collections.
\vspace{2mm}
Despite these advances, existing 3D geometric foundation models still suffer from limited accuracy and scalability. Their feed-forward predictions provide strong global priors, but the estimated camera poses and scales are often only approximately correct, which limits their use in high-fidelity applications such as Neural Radiance Fields (NeRF)~\cite{mildenhall2021nerf} and other view-synthesis pipelines. A key reason is that many models are trained with Structure from Motion (SfM)-derived supervision, where poses and geometry are generated by tools such as COLMAP~\cite{schonberger2016structure} rather than measured ground truth, thereby transferring the noise and bias of the reconstruction pipeline to the learned predictions. Meanwhile, GPU memory constraints make it difficult to process long sequences or large image collections in a single forward pass. Recent VGGT-based methods~\cite{maggio2025vggt, maggio2025vggt-slam2, xiong2026vggt, deng2025vggtlongchunkitloop} address this by splitting input into chunks and aligning chunk-level predictions with simple $\mathrm{SE}(3)$ or $\mathrm{Sim}(3)$ transformations, but this stitching strategy provides limited cross-chunk constraints and can accumulate pose and scale errors. Other approaches based on test-time training, recurrent memory, or sequence-specific architectures~\cite{wang2025continuous,xie2026scal3rscalabletesttimetraining,zhang2026loger, deng2023nerf} improve scalability, but often require backbone retraining or assume sequential inputs, reducing their flexibility in unordered image collections.
\vspace{2mm}

Compared with learning-based feed-forward predictions, classical SfM pipelines can often achieve higher pose accuracy by explicitly establishing correspondences and refining geometry with optimization. Their accuracy mainly comes from explicit geometric constraints, such as triangulation and bundle adjustment (BA). Existing SfM pipelines are commonly divided into incremental and global paradigms. Incremental SfM, such as COLMAP~\cite{schonberger2016structure}, registers images one by one and repeatedly applies BA, making it robust to noisy pairwise geometry but computationally expensive and sensitive to registration order. Global SfM, such as GLOMAP~\cite{pan2024glomap}, instead builds a pose graph from pairwise relative motions and estimates all camera poses jointly through rotation and translation averaging followed by BA, making it substantially faster than incremental reconstruction. However, its performance still depends on the quality of the pose graph, and unreliable correspondences or outlier relative poses can make the reconstruction brittle in challenging scenarios.
\vspace{2mm}

These observations suggest a natural direction: using 3D foundation models to predict multi-view correspondences together with relative motion and dense geometry, and then adopting global SfM principles to optimize the reconstruction. This combination is well-suited to the limitations of both paradigms. The foundation model provides robust geometric initialization, including camera poses, point maps, and confidence estimates, which enables the construction of a reliable pose graph. More importantly, its intermediate features encode cross-view geometric and structural cues, making them effective for correspondence prediction and reducing outlier matches. Given these correspondences and the resulting pose graph, global pose estimation and bundle adjustment can jointly refine camera poses and scene structure, substantially improving reconstruction accuracy.
\vspace{2mm}

Specifically, we build our framework on top of Pi3X~\cite{wang2025pi3} and introduce a dense matching head that predicts image warps and confidence between selected frames and other views. The dense warps are converted into sparse but reliable multi-view tracks, providing explicit correspondence constraints for global optimization. To efficiently select keyframes and associate long sequences, we further propose a keyframe-based sliding-window strategy that leverages the predicted local point maps, camera poses, and confidence maps. Instead of merely stitching independent chunks, our method uses overlapping windows to propagate tracks and relative poses across the full sequence, while performing subsequent optimization at the frame level. These local predictions and track associations are first converted into an initial global pose graph. We then perform motion averaging (including rotation and translation averaging) followed by bundle adjustment to reduce scale inconsistencies, refine camera poses, and recover dense geometry from the optimized reconstruction.
Extensive experiments across various benchmarks demonstrate the effectiveness of our pipeline. Our method improves novel-view synthesis by 2--3 dB PSNR over feed-forward baselines and about 1 dB over COLMAP-based poses. On KITTI, it reduces trajectory RMSE by 10\%--50\% compared with recent streaming methods~\cite{xie2026scal3rscalabletesttimetraining,chen2026geometric,zhang2026loger}. On ETH3D, it substantially improves rotation accuracy and nearly doubles translation accuracy over the latest learning-based SfM baseline~\cite{wang2025amb3r}.
\vspace{2mm}

Our main contributions are summarized as follows:
\begin{itemize}[leftmargin=1.2em, itemsep=0.2em, topsep=0.2em]
    \item We propose a foundation-model-guided framework that turns feed-forward 3D predictions into optimizable geometric constraints, combining robust learned priors with global SfM refinement.

    \item We introduce a dense warping module and a keyframe-based sliding-window association strategy to extract reliable feature tracks across long sequences and unordered image collections.

    \item Extensive experiments on indoor, outdoor, large-scale driving, and unordered SfM benchmarks demonstrate that our method significantly improves geometric accuracy over previous methods.
\end{itemize}

\input{figures/pipeline.tex}

%% file: figures/pipeline.tex
\begin{figure}
    \centering
    \includegraphics[width=1.00\linewidth]{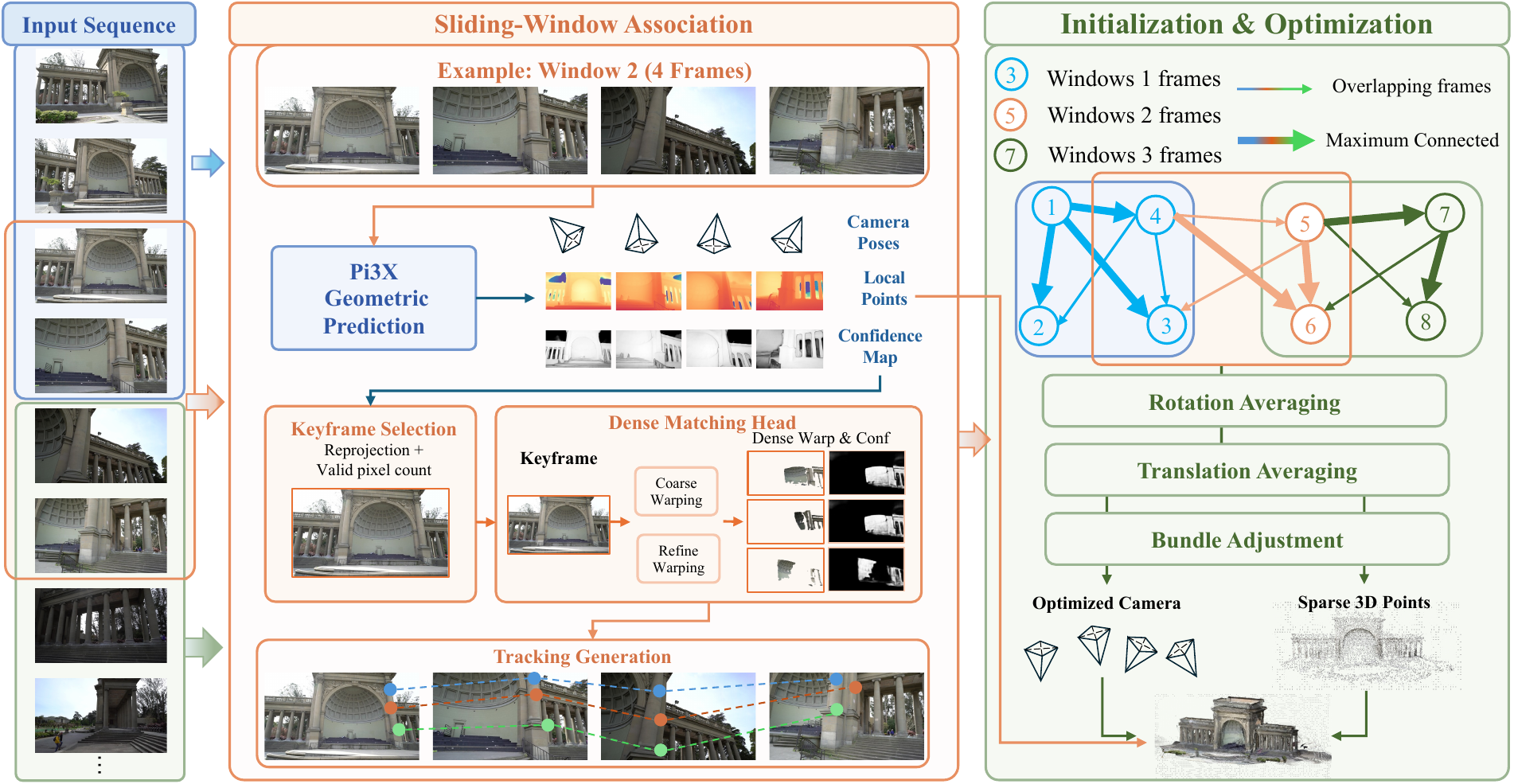}
    \caption{\textbf{Overview of the proposed reconstruction framework.} Given an ordered image sequence or a retrieval-based pseudo-sequence, we first predict geometric priors and dense warps within windows. Keyframes are selected according to reprojection coverage, and dense warps are converted into sparse multi-view tracks. These tracks are then merged into a global association graph for rotation averaging, translation averaging, bundle adjustment, and dense reconstruction.}
    \label{fig:pipeline}
    \vspace{-4mm}
\end{figure}

%% file: sections/02_related_works.tex
\vspace{-2mm}
\section{Related Works}
\vspace{-2mm}
\subsection{Structure-from-Motion and Geometric Optimization}
\vspace{-2mm}
Structure-from-Motion (SfM) is a fundamental problem in computer vision. Incremental SfM, such as COLMAP~\cite{schoenberger2016sfm}, progressively registers images starting from an initial image pair. In contrast, global SfM methods, such as GLOMAP~\cite{pan2024glomap}, estimate pairwise relative poses and jointly recover all camera poses via rotation averaging~\cite{hartley2013rotation, martinec2007robust,li2022rago}, translation averaging~\cite{govindu2001combining, wilson2014robust}, or global positioning~\cite{pan2024glomap}, followed by bundle adjustment~\cite{triggs1999bundle}. Despite their high reconstruction accuracy,
these optimization-based methods remain limited by the cost and reliability of feature matching, especially in large-scale scenes, weakly textured regions, repetitive structures, or forward-moving trajectories. Recent differentiable SfM~\cite{tang2018ba, wei2020deepsfm,teed2018deepv2d, teed2021droid,brachmann2024acezero,smith2024flowmap,wang2024vggsfm} methods further explore learnable reconstruction by enforcing geometric constraints and minimizing reprojection or photometric errors.

\vspace{-2mm}
\subsection{3D Foundation Models}
\vspace{-2mm}
3D foundation models have recently emerged as a feed-forward paradigm for 3D reconstruction. DUSt3R~\cite{wang2024dust3r} predicts dense scene coordinate maps for a pair of input views, while VGGT~\cite{wang2025vggt}, Pi3~\cite{wang2025pi3}, and DA3~\cite{depthanything3} predict camera, depth, and other 3D attributes with a single transformer network. These models are efficient and robust, and can solve many 3D vision tasks in an end-to-end manner with minimal hand-crafted assumptions. However, their geometric accuracy is often bounded by SfM-derived supervision such as COLMAP poses, and scaling them to large image collections remains challenging. Recent methods such as VGGT-Long~\cite{deng2025vggtlongchunkitloop} and VGGT-SLAM~\cite{maggio2025vggt} process long sequences by splitting them into chunks and estimating transformations between chunks, but such strategies underuse intra-chunk geometric relations and can accumulate errors across chunk. Some methods~\cite{li2025wint3r, wu2025point3r, cheng2026longstream, wang2025continuous, xiong2026vggt} maintain a memory bank to extend reconstruction to larger scenes, but the accumulated memory updates can still lead to substantial drift as the scene scale increases. Other approaches improve scalability through test-time training, recurrent memory, or sequence-specific architectures~\cite{xie2026scal3rscalabletesttimetraining,zhang2026loger,chen2026geometric}, but often require backbone retraining or assume sequential inputs.

\vspace{-2mm}
\subsection{Feature Matching and Dense Correspondence}
\vspace{-2mm}
Feature matching is crucial for both SfM and SLAM. Classical sparse pipelines detect keypoints and then establish correspondences by matching local descriptors, while learned matchers such as SuperGlue~\cite{sarlin2020superglue} and LightGlue~\cite{lindenberger2023lightglue} improve matching quality through attention-based reasoning. Detector-free methods, including LoFTR~\cite{sun2021loftr}, DKM~\cite{edstedt2023dkm}, and RoMa~\cite{edstedt2024roma}, move toward semi-dense or dense correspondence estimation and are more robust in weakly textured regions. Recently, RoMa v2~\cite{edstedt2025roma} further formulates two-view dense matching as image warping, achieving strong matching accuracy. Feature matching has also been explored in 3D foundation models: MASt3R~\cite{murai2025mast3r} predicts dense descriptors on top of DUSt3R, and VGGT~\cite{wang2025vggt} tracks query points across views. Inspired by these works, we construct multi-view correspondences through dense warping and derive accurate feature tracks for subsequent global pose optimization.

%% file: sections/03_method.tex
\section{Method}

Fig.~\ref{fig:pipeline} illustrates the pipeline of our method. Given an image collection, our goal is to estimate globally consistent camera poses and recover scene geometry. 

Sec.~\ref{sec:architecture} introduces our network architecture based on Pi3X: images are processed by a frozen 3D foundation model to obtain coarse geometric priors, including camera poses, local point maps, and confidence maps. For predicting multi-view feature matching, we further introduce a dense matching head that predicts image warps and confidence between selected keyframes and other frames. Sec.~\ref{sec:sliding_window} describes the sliding-window association strategy, which selects keyframes, converts dense warps into sparse multi-view tracks, and connects local windows into a  pose graph. Sec.~\ref{sec:optimization} presents the global optimization stage, including motion averaging, bundle adjustment, and dense reconstruction.

\subsection{Architecture}
\label{sec:architecture}
We build our model on top of Pi3X~\citep{wang2025pi3}, which adopts a unified transformer backbone with multiple prediction heads $f$ to infer geometric information from an image set ${\mathcal{I} = \left \{  I_i\in\mathbb{R}^{3\times H\times W} \right\}_{i=1}^{N} }$:
\vspace{-3mm}
\begin{equation}
    \label{sec3:eq:pi3x}
    f(\left\{ I_i \right\}_{i=1}^N) = \left\{ \mathbf{T}_i, \mathbf{X}_i, \mathbf{C}_i, m_i \right\}
\end{equation}
where for input image $I_i$, $\mathbf{T}_i\in SE(3) \subset \mathbb{R}^{4\times 4}$ denotes the camera pose, $\mathbf{X}_i \in \mathbb{R}^{3\times H\times W}$ is the 3D point map in the coordinate system of image $I_i$, $\mathbf{C}_i\in \mathbb{R}^{H\times W}$ is the corresponding confidence map, and $m_i$ denotes the approximate metric scale. 

We follow the permutation-equivariant design of Pi3X, which removes the need for a designated reference view and enables more flexible multi-view inputs.

To equip the model with dense matching capability, VGGT~\citep{wang2025vggt} takes a query point in a reference image as input and employs a tracking head to predict its corresponding points across all other images. Despite its effectiveness, the memory cost grows rapidly with the number of query points, and the resulting point-wise correspondences are less spatially connected than dense matching. Other methods, such as MASt3R~\citep{murai2025mast3r}, predict dense feature descriptors and confidence maps for every pixel, followed by nearest-neighbor search for correspondence estimation. However, dense descriptor prediction is computationally expensive and can be less accurate. Inspired by RoMa V2~\citep{edstedt2025roma}, we formulate dense matching as image warping, which maps 2D pixels from one image to another through a two-stage process consisting of coarse matching and subsequent refinement. 

Visualizations of the warping results are available in the supplementary material. 
Specifically, let $\mathbf{H}\in\mathbb{R}^{N\times L\times C}$ denote the output feature of the Pi3X backbone, where $N$ is the number of input images, $L$ is the token length, and $C$ is the feature dimension. Given a reference image $I_a$, we treat the remaining images as target views $\mathcal{B}=\{1,\cdots,N\}\setminus\{a\}$. The coarse matching results are predicted by a transformer decoder ($\mathrm{Dec}_{\mathrm{match}}$) followed by a Dense Prediction Transformer ($\mathrm{DPT}_{\mathrm{match}}$):
\begin{equation}
\label{equ:matching}
    \left(
    \mathbf{W}^{a\rightarrow \mathcal{B}},
    \mathbf{p}^{a\rightarrow \mathcal{B}}
    \right)
    =
    \mathrm{DPT}_{\mathrm{match}}
    \left(
    \mathrm{Dec}_{\mathrm{match}}(\mathbf{H}), a
    \right),
\end{equation}
where $\mathbf{W}^{a\rightarrow \mathcal{B}}\in\mathbb{R}^{(N-1)\times 2\times \frac{H}{4}\times \frac{W}{4}}$ maps pixels from the reference image to each target image, and $\mathbf{p}^{a\rightarrow \mathcal{B}}\in\mathbb{R}^{(N-1)\times \frac{H}{4}\times \frac{W}{4}}$ denotes the corresponding confidence. Both outputs are predicted at a stride of 4 and then refined to full resolution. A refinement module takes the coarse matching results and the original image as input, and progressively predicts residual warps $\Delta W$ and confidence residuals $\Delta p$ at strides $\left\{4, 2, 1\right\}$. More details are provided in the supplementary material.

\noindent
\textbf{Loss Function and Training.} We freeze the encoder, decoder, and existing prediction heads, and fine-tune only the matching head. Since Pi3X is permutation-equivariant, the reference view can be chosen flexibly during inference, which is important for our subsequence-based pipeline. During training, we simply use the first image as the reference view and supervise its matches to the remaining target images with ground-truth 2D--2D warp. The ground-truth warp is generated by projecting the ground-truth 3D point map with known camera poses, and the confidence supervision is derived from depth consistency. The total loss is the weighted sum of the negative log-likelihood of the best matching patch, the warping loss, and the confidence loss:
\vspace{-1mm}
\begin{equation}
    \mathcal{L} = \sum_{k = 2}^{N}  \lambda_1 \mathcal{L}^{k}_{\mathrm{NLL}} + \lambda_2 \mathcal{L}^{k}_{\mathrm{warp}} + \lambda_3 \mathcal{L}^{k}_{\mathrm{conf}}
\end{equation}
where $\mathcal{L}^{k}_{\mathrm{NLL}}$ encourages high cosine similarity between the features of matching patches. Please refer to the supplementary material for more details.

\subsection{Sliding-window Sequence Association}
\label{sec:sliding_window}

Eq.~\ref{sec3:eq:pi3x} provides a robust geometric prior such as camera pose and point map, while Eq.~\ref{equ:matching} gives image warps between frames. They can connect a pose graph through sparse but reliable multi-view tracks.

\noindent
\textbf{Sequence and Window Construction.}
Our method operates on an image sequence $\mathcal{S}=\{I_i\}_{i=1}^{M}$. For data with explicit temporal order, we use the original order. For unordered image sets, we build a pseudo-sequence with an image retrieval model such as SALAD~\citep{Izquierdo_CVPR_2024_SALAD}. This fits our framework because the network does not require strictly ordered frames. Any frame in the window with sufficient visual overlap can be the reference view.
We process the sequence with sliding windows, where each window contains $N$ frames and is shifted by $N/2$ frames along the sequence. The resulting half-window overlap provides shared frames between adjacent windows, enabling correspondences and relative poses to be propagated progressively across the full sequence.

\noindent
\textbf{Keyframe Selection.} Following the principle of classical SLAM systems, we use a small number of keyframes to establish local mappings, while the remaining frames are associated through the matches to the keyframes. For each window, we initialize its first frame as a keyframe if there is no keyframe. We then run Eq.~\ref{sec3:eq:pi3x} to obtain the camera pose $\mathbf{T}_t$, the point map $\mathbf{X}_t$ applying the predicted metric scale, and the confidence map $\mathbf{C}_t$ for every frame. For a candidate frame $I_t$, we reproject its point map onto each existing keyframe $I_r \in \mathcal{K}$ and count the number of valid projected pixels:

\begin{equation}
    n_t =
    \max_{I_r \in \mathcal{K}}
    \sum_{\mathbf{u}}
    \mathbf{1}
    \left[
    \pi\left(\mathbf{T}_{t\rightarrow r}\bar{\mathbf{X}}_t(\mathbf{u})\right)
    \in \mathcal{D},
    \ z_{t\rightarrow r}(\mathbf{u}) > 0,
    \ \mathbf{C}_t(\mathbf{u}) > \tau_c
    \right],
\end{equation}
where $\mathcal{K}$ is the current keyframe set, $\bar{\mathbf{X}}_t(\mathbf{u})$ denotes the homogeneous 3D point at pixel $\mathbf{u}$, $\mathbf{T}_{t\rightarrow r}$ is the relative transform from $I_t$ to $I_r$, $\pi(\cdot)$ is the projection function, $\mathcal{D}$ is the image domain, $z_{t\rightarrow r}(\mathbf{u})$ is the projected depth, and $\tau_c$ is the depth confidence threshold. If $n_t<\tau_{\mathrm{proj}}$, we add $I_t$ to the keyframe set. This process is repeated until all frames in the window have been visited. We additionally ensure that at least one keyframe lies in the overlapping half of the window to maintain connectivity between consecutive windows.

\noindent
\textbf{Track-based Global Initialization.}
After keyframe selection, we use the matching head to predict dense warps from each keyframe to all frames in the same window. 

This step only requires a single forward pass of the frozen backbone to extract features, after which the backbone features are reused and kept fixed for keyframe-to-frame matching, resulting in moderate memory overhead.
We sample $K$ high confidence pixels from the keyframe point map, and pick the corresponding pixel on the dense warp map of other frames to build the track. The tracking confidence is obtained by combining the keyframe point confidence and the predicted warp confidence. This produces sparse yet reliable multi-view tracks anchored at keyframes.

The local tracks from the overlapping windows are merged into a pose graph. Each frame is treated as a node, and an edge is added between two frames if they share valid tracking correspondences. The edge weight is defined by the number of valid tracks between the two frames. Since the matching results and local geometry predictions are produced jointly within each window, each valid edge is associated with a relative pose. For edges from different windows, we temporarily use the metric scale predicted by Pi3X~\cite{wang2025pi3} to bring their relative transformation into global space. We then compute a maximum spanning tree over the graph and initialize the camera pose with relative poses on the spanning tree. The initialized poses and the associated sparse tracks are then passed to the subsequent optimization stage, where remaining scale inconsistencies and coarse poses are further refined.

This design differs from window-stitching pipelines~\citep{deng2025vggtlongchunkitloop, maggio2025vggt}, which mainly merge independent chunk-level predictions. By using frames as the basic association and optimization unit, our method not only connects different windows through overlapping keyframes, but also enables pose refinement of all input images, leading to more accurate and globally consistent results.

\subsection{Optimization}
\label{sec:optimization}

The global initialization provides each frame with a reasonable pose estimate, while its  scale, inherited from the foundation model, remains approximate and may vary across windows. Moreover, each frame can be connected to multiple keyframes from overlapping windows, resulting in redundant but not always consistent relative pose constraints. We therefore perform global optimization to consolidate local predictions into a coherent reconstruction.

\noindent
\textbf{Motion Averaging.}
We first estimate globally consistent camera poses from the pose graph. Inspired by GLOMAP~\cite{pan2024glomap}, we decompose this step into rotation averaging followed by translation averaging. Given the relative rotations associated with the graph edges, we perform robust rotation averaging~\cite{hartley2013rotation} to obtain a globally consistent rotation $\mathbf{R}_i$ for each frame. 
With rotations fixed, we estimate camera centers and sparse 3D points using multi-view ray consistency. For a track observation $\mathbf{u}_{ij}$ of point $j$ in frame $i$, we convert it into a normalized camera ray $\mathbf{v}_{ij}$. Instead of optimizing pairwise translation scales, we require the 3D point to lie on the corresponding world-frame viewing ray:
\begin{equation}
    \min_{\{\mathbf{c}_i\}, \{\mathbf{X}_j\}, \{d_{ij}\}}
    \sum_{(i,j)\in\mathcal{O}}
    \omega_{ij}
    \rho
    \left(
    \left\|
    \mathbf{X}_j -
    \left(
    \mathbf{c}_i + d_{ij}\mathbf{R}_i^{\top}\mathbf{v}_{ij}
    \right)
    \right\|_2^2
    \right),
\end{equation}
where $\mathcal{O}$ is the set of valid track observations, $\mathbf{c}_i$ is the camera center, $\mathbf{X}_j$ is the 3D point, $d_{ij}$ is the depth of point $j$ along the viewing ray of frame $i$, $\omega_{ij}$ is the tracking confidence, and $\rho(\cdot)$ is a robust loss. This formulation is robust to inconsistent local translation scales and recovers globally compatible camera centers and sparse points from multi-view tracks. The resulting poses provide a stable initialization for the subsequent bundle adjustment.

\noindent
\textbf{Bundle Adjustment.}
Starting from the global initialization, we further refine camera poses, sparse points, and camera intrinsics by minimizing reprojection errors over all valid track observations:
\begin{equation}
    \min_{\{\mathbf{T}_i\}, \{\mathbf{X}_j\}, \{\mathbf{K}_i\}, \{\boldsymbol{\delta}_i\}}
    \sum_{(i,j)\in\mathcal{O}}
    \omega_{ij}
    \rho
    \left(
    \left\|
    \pi\left(\mathbf{K}_i, \boldsymbol{\delta}_i, \mathbf{T}_i, \mathbf{X}_j\right)
    -
    \mathbf{u}_{ij}
    \right\|_2^2
    \right),
\end{equation}
where $\mathbf{T}_i$ is the camera pose, $\mathbf{K}_i$ is the camera intrinsic matrix, $\boldsymbol{\delta}_i$ denotes the distortion parameters, and $\pi(\cdot)$ is the projection function with distortion. When camera calibration is available, we keep $\mathbf{K}_i$ and $\boldsymbol{\delta}_i$ fixed; otherwise, they are jointly optimized with poses and points. The confidence $\omega_{ij}$ down-weights uncertain matches, while the robust loss suppresses outliers from incorrect tracks and occlusions. This two-stage optimization first resolves large-scale inconsistencies and then performs accurate reprojection-based refinement, allowing poses to be optimized at the frame level rather than merely stitched at window boundaries.

\noindent
\textbf{Dense Reconstruction.}
During initialization, we also store the predicted depth and confidence map of each frame. After optimization, the sparse 3D tracks provide reliable depth samples at their corresponding pixels. For each keyframe, we estimate a scale factor between the predicted dense depth and the optimized sparse depths using RANSAC, and rescale the dense depth accordingly. The rescaled depth is then back-projected with the optimized intrinsics and distortion parameters to obtain a dense point map in the camera coordinate system. Finally, all dense point maps are transformed by the optimized poses and fused into a globally aligned dense point cloud.

%% file: sections/04_experiment.tex
\section{Experiments}

\subsection{Experimental Setup}

\noindent
\textbf{Datasets.}
We evaluate our method on diverse indoor, outdoor, unordered SfM, and large-scale driving scenes, including 19 Tanks and Temples (T\&T) scenes~\cite{knapitsch2017tanks}, 9 TUM RGB-D sequences~\cite{sturm12iros}, 13 ETH3D scenes~\cite{schoeps2017eth3d}, and 11 KITTI driving sequences~\cite{Geiger2012kitti}. T\&T, TUM RGB-D, and KITTI are evaluated as streaming or sequential inputs, while ETH3D is treated as an unordered SfM-style image collection. These datasets cover diverse scene scales, camera motions, and visual conditions.

\noindent
\textbf{Evaluation Metrics and Inference Settings.}
We mainly evaluate the camera pose accuracy. For T\&T, accurate ground-truth poses are not available, and using COLMAP poses as ground truth would be inappropriate since they are also estimated results~\citep{brachmann2024acezero}. We therefore follow the previous works~\citep{brachmann2024acezero,dengli2025sail} to evaluate pose quality through novel-view synthesis. For each method, we split the estimated camera pose into training/testing splits, and then train a Nerfacto model~\citep{nerfstudio} on training views and report PSNR on rendered testing views as an indirect measure of pose precision.
For TUM RGB-D and KITTI, ground-truth trajectories are available, so we align the estimated trajectories to the ground truth and report pose RMSE. For ETH3D, following common relative-pose evaluation protocols, we report RRA@5 and RTA@5, i.e., the proportion of camera pairs whose relative rotation and translation errors are below $5^\circ$.
By default, we use a sliding window size of $N=20$ with a stride of 10 frames. 

We use $N=120$ for T\&T Auditorium and Courtroom, and $N=200$ for the KITTI sequence 02 to stabilize Pi3X predictions while leaving the rest of the pipeline unchanged. 

Please refer to the supplementary for more details.

\subsection{Comparison with State-of-the-art Methods}

We compare our method with a broad set of reconstruction and SLAM baselines, including classical SfM methods COLMAP~\cite{schonberger2016structure}, GLOMAP~\cite{pan2024glomap}, feed-forward 3D foundation models such as DA3~\cite{depthanything3} and Pi3X~\cite{wang2025pi3}, streaming SLAM systems such as DROID-SLAM~\cite{teed2021droid}, and recent long-sequence or SfM-style learning-based methods, including VGGT-Long~\cite{deng2025vggtlongchunkitloop}, VGGT-SLAM~\cite{maggio2025vggt}, LingBot-Map~\cite{chen2026geometric}, LoGeR~\cite{zhang2026loger}, SCAL3R~\cite{xie2026scal3rscalabletesttimetraining}, SAILRecon~\cite{dengli2025sail}, MASt3R-SLAM~\cite{murai2025mast3r}, and AMB3R~\cite{wang2025amb3r}. Please refer to the supplementary for more baseline details and additional comparisons.

\input{figures/tntpsnr1.tex}
\input{tables/tnt.tex}

\noindent
\textbf{Tanks and Temples.}
Table~\ref{tab:tnt} reports novel-view synthesis results on T\&T, where higher PSNR indicates more accurate and consistent camera poses. Our method achieves the best average PSNR, outperforming both the classical SfM systems COLMAP~\cite{schonberger2016structure} and GLOMAP~\cite{pan2024glomap} and the learning-based reconstruction methods on average. Although GLOMAP obtains strong results on several scenes, its performance is less stable due to inaccurate edges and outliers in the pose graph. In contrast, our method benefits from the geometric priors of the foundation model, which provide robust initial poses, point maps for subsequent optimization. Compared with feed-forward methods such as DA3~\cite{depthanything3} and Pi3X~\cite{wang2025pi3}, our method achieves better reconstruction quality, suggesting that feed-forward global reasoning alone is insufficient for high-precision pose estimation. Recent streaming or post-optimization pipelines, including LingBot-Map~\cite{chen2026geometric}, LoGeR~\cite{zhang2026loger}, Scal3R~\cite{xie2026scal3rscalabletesttimetraining}, and SAIL-Recon~\cite{dengli2025sail}, also remain inferior on this benchmark. This reveals our motivation: 3D foundation models provide robust but coarse pose initialization and geometry-aware features, while classical global SfM can refine geometry but is sensitive to unreliable pose graphs. By converting dense warps into reliable multi-view tracks, our method bridges these two sides and enables stable global SfM-style optimization for high-fidelity reconstruction. The qualitative results in Fig.~\ref{fig:tntpsnr1} confirm this observation, where our rendered views are visually closer to the ground-truth images.

\input{tables/tum.tex}
\noindent
\textbf{TUM RGB-D.}
Table~\ref{tab:tum_pose} evaluates pose accuracy in small-scale RGB-D indoor sequences. Our method achieves the best average RMSE among these uncalibrated methods and remains competitive with recent SLAM and reconstruction systems. 
This benchmark contains short indoor trajectories with frequent rotations, limited baselines, and appearance variations, posing challenges for feature matching followed by geometric optimization.
Although our method requires explicit feature matching, it remains stable under these difficult motion settings. These results show that our approach is effective for compact indoor sequences where accurate local pose refinement is required.
\input{figures/kitti0009.tex}

\input{tables/kitti.tex}
\noindent
\textbf{KITTI.}
Table~\ref{tab:kitti_pose} evaluates long-sequence driving scenarios. Our method achieves the best average RMSE across all KITTI sequences. This benchmark is challenging because forward-facing vehicle motion often produces small parallax, weak triangulation, and unstable feature matching. Classical SfM methods~\cite{schonberger2016structure,pan2024glomap} are therefore prone to failure or high computational cost in such settings. Recent scalable or streaming reconstruction methods, including VGGT-Long~\cite{deng2025vggtlongchunkitloop}, Scal3R~\cite{xie2026scal3rscalabletesttimetraining}, LoGeR~\cite{zhang2026loger}, and LingBot-Map~\cite{chen2026geometric}, alleviate the memory issue with chunk-wise processing, test-time optimization, or streaming updates, but still accumulate trajectory errors over long driving sequences. In contrast, our sliding-window association converts dense warps into sparse tracks and propagates them across overlapping windows, enabling frame-level pose optimization over the full sequence. The qualitative trajectories in Fig.~\ref{fig:kitti0009} show the same trend: our method closely follows the ground truth on KITTI Odometry 00 and 09, while competing methods exhibit more visible drift or trajectory deformation.

\input{tables/eth3d.tex}
\noindent
\textbf{ETH3D.}
Table~\ref{tab:eth3d} reports results on unordered SfM-style image collections. We first compare with GLOMAP~\cite{pan2024glomap} using the same view graph connectivity. Although our optimization follows a similar global SfM formulation, GLOMAP drops noticeably in this setting because erroneous edges in the view graph can strongly affect rotation and translation averaging. In contrast, our method benefits from geometric foundation model priors and confidence-aware dense matching, which provide more reliable tracks for global optimization. Our method also significantly outperforms strong learning-based baselines such as AMB3R-SfM~\cite{wang2025amb3r}. Under the loose threshold of $5^\circ$, our method reaches saturated or near-saturated RRA@5 and RTA@5 on most scenes; under the stricter $1^\circ$ threshold, it still achieves substantially higher RRA@1 and RTA@1 than AMB3R, indicating more accurate relative poses for high-precision SfM evaluation.

\input{tables/abla.tex}

\subsection{Ablation Study}

We perform ablation studies on ETH3D with accurate ground-truth poses. Table~\ref{tab:abla} compares different variants using RRA@1 and RTA@1. The initialization achieves reasonable accuracy, showing that the system can connect different frames to a meaningful global structure. Motion Averaging further brings a modest improvement by making the initial poses more globally consistent. After complete optimization, our method improves both rotation and translation accuracy, confirming the importance of geometric bundle adjustment.

We further analyze the matching component. Using the coarse matching gives limited improvement over initialization, indicating that the refinement is important for accurate correspondence. Replacing our matching with the VGGT~\cite{wang2025vggt} tracking head or RoMaV2~\cite{edstedt2025roma} leads to poorer accuracy and lower efficiency. 

This is because VGGT tracks points from a single reference view, so multiple keyframes require repeated inference; moreover, the resulting tracks are less reliable than the points obtained from global dense warping. RoMaV2 only operates as a two-view matcher and does not exploit the shared multi-view geometric context provided by the foundation model. These results show that our dense matching head provides more accurate and efficient tracks for global pose optimization.

%% file: figures/tntpsnr1.tex
\begin{figure}
    \centering
    \includegraphics[width=1.00\linewidth]{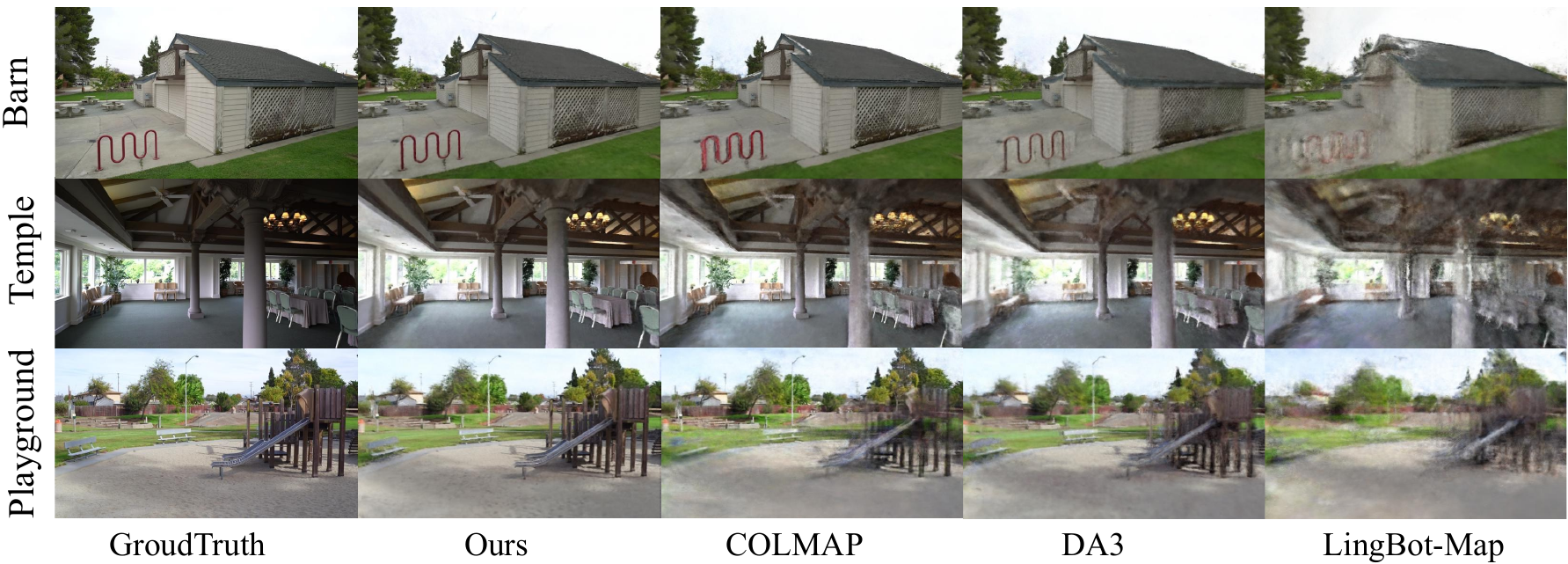}
    \caption{Qualitative comparison of novel-view synthesis on Tanks and Temples. Our method produces cleaner renderings with fewer pose-induced artifacts.}
\label{fig:tntpsnr1}
    \vspace{-4mm}
\end{figure}

%% file: tables/tnt.tex
\begin{table*}[t]
\centering
\caption{PSNR comparison on selected Tanks and Temples scenes. Best results are in bold.}
\label{tab:tnt}
\small
\setlength{\tabcolsep}{3.0pt}
\renewcommand{\arraystretch}{0.95}
\resizebox{\textwidth}{!}{
\begin{tabular}{lccccccccccccccc}
\toprule
Scene 
& Barn & Cat. & Chur. & Court. & C.room & Family & Franc. & Horse 
& Light. & M.room & Museum & Play. & Temple & Train & Avg. \\
\midrule
COLMAP~\citep{schonberger2016structure} 
& 24.09 & 17.04 & 18.14 & 18.04 & 18.25 & 19.40 & 21.80 & 19.47 
& 16.65 & 18.59 & 16.87 & 19.07 & 18.10 & \textbf{16.70} & 18.73 \\
GLOMAP~\citep{pan2024glomap} 
& 24.26 & 16.97 & \textbf{18.20} & \textbf{21.08} & 18.20 & 13.08 & 22.00 & 14.85 
& 17.76 & 19.56 & 11.81 & 15.72 & 18.23 & 12.35 & 17.43 \\
AMB3R~\cite{wang2025amb3r} 
& 20.08 & 15.45 & 15.98 & 15.64 & 15.58 & 19.50 & 19.19 & 19.08 
& 15.00 & 16.63 & 15.97 & 18.04 & 15.74 & 15.66 & 16.97 \\
Scal3R~\citep{xie2026scal3rscalabletesttimetraining}
& 20.00 & 14.86 & 15.08 & 18.20 & 13.28 & 17.21 & 16.08 & 16.58 
& 12.09 & 14.74 & 13.01 & 16.79 & 17.09 & 13.04 & 15.58 \\
DA3~\cite{depthanything3}
& 21.78 & 15.73 & 16.51 & OOM & 16.34 & 18.93 & 20.38 & 18.60 
& 17.43 & 17.78 & 15.60 & 19.01 & 17.21 & 15.88 & 17.78 \\
Pi3X~\citep{wang2025pi3} 
& 21.22 & 16.04 & 17.01 & 16.74 & 17.23 & 18.89 & 20.27 & 18.19 
& 17.34 & 17.63 & 15.56 & 18.20 & 16.83 & 15.57 & 17.62 \\
SAIL-R.~\citep{dengli2025sail} 
& 23.50 & 16.80 & 17.00 & 15.09 & 17.40 & \textbf{20.60} & 21.80 & \textbf{20.10} 
& 18.20 & 19.50 & 15.40 & 20.30 & 17.80 & 16.20 & 18.55 \\
LingBot~\cite{chen2026geometric}
& 18.54 & 14.51 & 15.44 & 14.43 & 15.65 & 17.60 & 19.01 & 17.17 
& 15.29 & 16.08 & 14.62 & 17.18 & 15.28 & 14.77 & 16.11 \\
Ours 
& \textbf{24.50} & \textbf{17.25} & 18.01 & 20.72 & \textbf{18.31} & 20.27 
& \textbf{22.01} & 19.79 & \textbf{18.76} & \textbf{19.68} & \textbf{17.43} 
& \textbf{21.68} & \textbf{18.82} & 16.60 & \textbf{19.56} \\
\bottomrule
\end{tabular}
}
\vspace{-2mm}
\end{table*}

%% file: tables/tum.tex
\begin{table*}[htbp]
\vspace{-2mm}
\centering
\caption{Pose estimation results on TUM RGB-D sequences. We report trajectory RMSE (cm) $\downarrow$.}
\label{tab:tum_pose}
\renewcommand{\arraystretch}{0.6}
\resizebox{0.99\textwidth}{!}{
\begin{tabular}{l|ccccccccc|c}
\toprule
Method & 360 & desk & desk2 & floor & plant & room & rpy & teddy & xyz & Avg. \\
\midrule
DROID-SLAM \citep{teed2021droid}
& 20.2 & 3.2 & 9.1 & 6.4 & 4.5 & 91.8 & 5.6 & 4.5 & 1.2 & 15.8 \\
MASt3R-SLAM \citep{murai2025mast3r}
& 7.0 & 3.5 & 5.5 & 5.6 & 3.5 & 11.8 & 4.1 & 11.4 & 2.0 & 6.0 \\
VGGT-SLAM~\cite{maggio2025vggt}
& 7.1 & 2.5 & 4.0 & 14.1 & 2.3 & 10.2 & 3.0 & 3.4 & 1.4 & 5.3 \\
AMB3R \citep{wang2025amb3r}
& \textbf{4.6} & 1.9 & 2.8 & 3.2 & 2.9 & 5.8 & 2.3 & 3.7 & 1.1 & \textbf{3.2} \\
LoGeR~\cite{zhang2026loger}
& 10.6 & 3.2 & 4.6 & 10.3 & 4.6 & 9.8 & 3.0 & 8.4 & 2.2 & 6.3 \\
Scal3R \cite{xie2026scal3rscalabletesttimetraining}
& 6.6 & 5.5 & 2.9 & 18.9 & 4.5 & 11.0 & 3.2 & 9.5 & 5.0 & 7.4 \\
VGGT-SLAM 2.0 \citep{maggio2025vggt-slam2} 
& 5.0 & 2.5 & 2.9 & 10.2 & 2.6 & 6.3 & 2.6 & 3.8 & 1.4 & 4.1 \\
SAILRecon \citep{dengli2025sail}
& 7.0 & 2.4 & 4.2 & 10.7 & 3.1 & 11.3 & \textbf{2.0} & 3.7 & 1.2 & 5.1 \\
LingBot-Map \citep{chen2026geometric}
& 6.3 & 2.8 & 4.4 & 5.9 & 4.4 & 9.2 & 2.4 & 3.6 & 1.1 & 4.4 \\
Ours 
& 8.1 & \textbf{1.8} & \textbf{2.6} & \textbf{3.0} & \textbf{1.8} & \textbf{5.0} & 2.2 & 3.7 & \textbf{0.9} & \textbf{3.2} \\
\bottomrule
\end{tabular}
}
\end{table*}
\vspace{-2mm}

%% file: figures/kitti0009.tex
\begin{figure}
    \centering
    \includegraphics[width=1.00\linewidth]{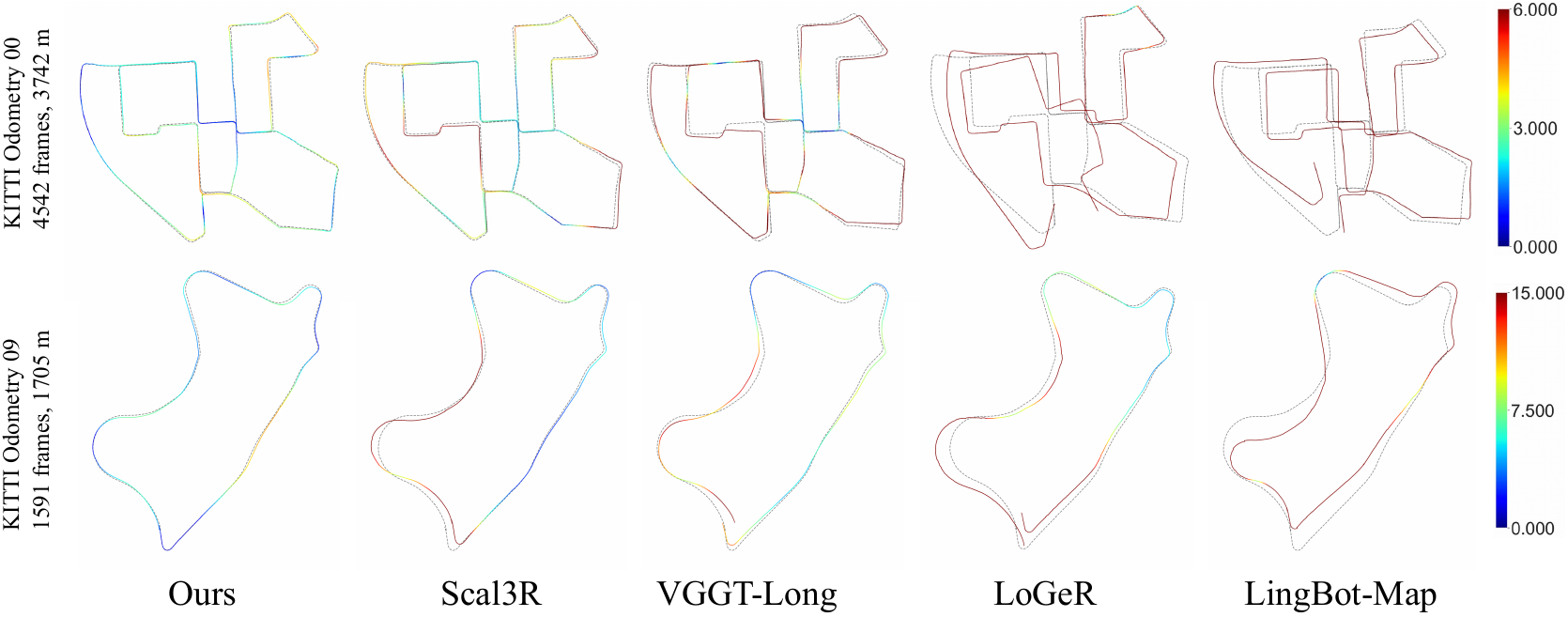}
    \caption{Camera trajectory comparison on KITTI Odometry. Our method better preserves the global trajectory shape and reduces drift on long driving sequences.}
    \label{fig:kitti0009}
    \vspace{-3mm}
\end{figure}

%% file: tables/kitti.tex
\begin{table*}[htbp]
\centering
\vspace{-4mm}
\caption{Pose estimation results on KITTI sequences. We report trajectory RMSE $(m) \downarrow$.}
\label{tab:kitti_pose}
\resizebox{\textwidth}{!}{
\begin{tabular}{lcccccccccccc}
\toprule
Method 
& {00} & {01} & {02} & {03} & {04} & {05} & {06} & {07} & {08} & {09} & {10} & {Avg.} \\
\small Frames / km
& \small 4542 / 3.7 & \small 1101 / 2.5 & \small 4661 / 5.1 & \small 801 / 0.6 & \small 271 / 0.4 
& \small 2761 / 2.2 & \small 1101 / 1.2 & \small 1101 / 0.7 & \small 4071 / 3.2 
& \small 1591 / 1.7 & \small 1201 / 0.9 & \\
\midrule
DROID-SLAM \citep{teed2021droid}
& 92.10 & 344.60 & 107.61 & 2.38 & 1.00 & 118.50 & 62.47 & 21.78 & 161.60 & 72.32 & 118.70 & 100.28 \\
VGGT-Long \citep{deng2025vggtlongchunkitloop}
& 8.64 & 61.21 & 52.72 & 8.78 & 4.20 & 9.88 & 4.67 & 2.66 & 72.98 & 31.84 & 27.71 & 25.94 \\
VGGT-SLAM 2.0 \citep{maggio2025vggt-slam2}
& TL & 163.65 & TL & 50.04 & 19.38 & 159.58 & 46.35 & 57.80 & TL & 167.96 & 76.99 & 92.72 \\
MASt3R-SLAM \citep{murai2025mast3r}
& OOM & 530.37 & OOM & 18.87 & 88.99 & 159.43 & 92.00 & OOM & 263.75 & TL & 153.07 & 186.64 \\
AMB3R~\cite{wang2025amb3r} 
& 167.38 & 276.12 & 157.07 & 22.06 & 6.91 & 149.11 & 55.94 & 37.25 & 85.68 & 134.73 & 52.15 & 104.04 \\
LingBot-Map \citep{chen2026geometric}
& 27.17 & 70.94 & 112.02 & 2.02 & 1.36 & 26.14 & 16.61 & 10.48 & \textbf{23.82} & 17.84 & 6.48 & 28.63 \\
LoGeR \cite{zhang2026loger}
& 30.47 & 47.91 & 36.32 & 5.38 & 1.95 & 26.34 & 6.60 & 5.55 & 24.41 & 10.12 & 10.11 & 18.65 \\
Scal3R \citep{xie2026scal3rscalabletesttimetraining}
& 4.30 & \textbf{45.29} & 42.06 & 3.36 & 1.74 & 3.30 & \textbf{2.49} & 2.03 & 36.69 & 12.32 & \textbf{6.46} & 14.55 \\
Ours 
& \textbf{2.77} & 53.12 & \textbf{30.10} & \textbf{1.53} & \textbf{0.75} & \textbf{2.80} 
& 2.67 & \textbf{1.77} & 32.76 & \textbf{5.51} & 11.51 & \textbf{13.21} \\
\bottomrule
\end{tabular}
}
\end{table*}
\vspace{-2mm}

%% file: tables/eth3d.tex
\begin{table}[t]
\vspace{-2mm}
\centering
\caption{Pose estimation accuracy comparison on ETH3D. Baselines are evaluated at loose thresholds (5$^\circ$). We report strict accuracy (1$^\circ$) for AMB3R and our method, and also provide our method's performance at loose thresholds (5$^\circ$) for comprehensive comparison. Best results are in bold.}
\resizebox{\textwidth}{!}{
\begin{tabular}{l|cccccccccccccc|cccc}
\toprule
\multirow{3}{*}{Scenes} 
& \multicolumn{14}{c|}{ Threshold @5} 
& \multicolumn{4}{c}{ Threshold @1} \\
\cmidrule(lr){2-15} \cmidrule(lr){16-19}
& \multicolumn{2}{c}{COLMAP \citep{schonberger2016structure}} 
& \multicolumn{2}{c}{GLOMAP~\cite{pan2024glomap}} 
& \multicolumn{2}{c}{VGGSfM \citep{wang2024vggsfm}} 
& \multicolumn{2}{c}{DF-SfM \citep{he2024detector}} 
& \multicolumn{2}{c}{MASt3R-SfM \citep{duisterhof2024mast3r}} 
& \multicolumn{2}{c}{AMB3R \citep{wang2025amb3r}} 
& \multicolumn{2}{c|}{\textbf{Ours}} 
& \multicolumn{2}{c}{AMB3R \citep{wang2025amb3r}} 
& \multicolumn{2}{c}{\textbf{Ours}} \\
\cmidrule(lr){2-3} \cmidrule(lr){4-5} \cmidrule(lr){6-7} 
\cmidrule(lr){8-9} \cmidrule(lr){10-11} \cmidrule(lr){12-13}
\cmidrule(lr){14-15} \cmidrule(lr){16-17} \cmidrule(lr){18-19}
& R@5 & T@5 & R@5 & T@5 & R@5 & T@5 & R@5 & T@5 & R@5 & T@5 & R@5 & T@5 & R@5 & T@5 & R@1 & T@1 & R@1 & T@1 \\
\midrule
courtyard     
& 56.3 & 60.0 & 30.33 & 26.52 & 50.5 & 51.2 & 80.7 & 74.8 & 89.8 & 64.4 & \textbf{100.0} & 96.5 & \textbf{100.0} & \textbf{97.0} & 86.57 & 50.28 & \textbf{100.0} & \textbf{73.89} \\
deli. area 
& 34.0 & 28.1 & 17.77 & 8.37 & 22.0 & 19.6 & 82.5 & \textbf{82.0} & 83.1 & 81.8 & 91.0 & 76.6 & \textbf{100.0} & 81.9 & 69.63 & 15.70 & \textbf{83.37} & \textbf{66.48} \\
electro       
& 53.3 & 48.5 & 57.23 & 50.12 & 79.9 & 58.6 & 82.8 & 81.2 & \textbf{100.0} & 95.5 & 95.6 & 81.2 & \textbf{100.0} & \textbf{85.3} & \textbf{80.05} & 36.74 & 72.05 & \textbf{55.95} \\
facade        
& 92.2 & 90.0 & 55.06 & 49.26 & 57.5 & 48.7 & 80.9 & 82.6 & 74.3 & 75.3 & \textbf{100.0} & 95.4 & \textbf{100.0} & \textbf{96.8} & 86.81 & 50.12 & \textbf{94.25} & \textbf{81.86} \\
kicker        
& 87.3 & 86.2 & 80.65 & 49.74 & \textbf{100.0} & 97.8 & 93.5 & 91.0 & \textbf{100.0} & \textbf{100.0} & \textbf{100.0} & 99.2 & \textbf{100.0} & 95.7 & \textbf{95.42} & 80.54 & 94.38 & \textbf{86.47} \\
meadow        
& 0.9 & 0.9 & 28.00 & 7.11 & \textbf{100.0} & \textbf{96.2} & 56.2 & 58.1 & 58.1 & 58.1 & \textbf{100.0} & 95.2 & \textbf{100.0} & 92.0 & \textbf{80.44} & 40.00 & 78.67 & \textbf{68.44} \\
office        
& 36.9 & 32.3 & 34.91 & 31.07 & 64.9 & 42.1 & 71.1 & 54.5 & \textbf{100.0} & \textbf{98.5} & \textbf{100.0} & 53.9 & \textbf{100.0} & 63.9 & \textbf{92.31} & 18.79 & 77.81 & \textbf{33.43} \\
pipes         
& 30.8 & 28.6 & 74.49 & 65.31 & \textbf{100.0} & 97.8 & 72.5 & 61.5 & \textbf{100.0} & \textbf{100.0} & \textbf{100.0} & 87.9 & \textbf{100.0} & 92.9 & 67.35 & 30.61 & \textbf{100.0} & \textbf{73.47} \\
p.ground    
& 17.2 & 18.1 & 75.90 & 28.12 & 37.3 & 40.8 & 70.5 & 70.1 & \textbf{100.0} & 93.6 & 98.7 & 62.2 & \textbf{100.0} & \textbf{95.4} & 69.94 & 19.74 & \textbf{90.03} & \textbf{68.01} \\
relief        
& 16.8 & 16.8 & 81.89 & 78.98 & 59.6 & 57.9 & 32.9 & 32.9 & 34.2 & 40.2 & \textbf{100.0} & 90.1 & \textbf{100.0} & \textbf{96.6} & 68.16 & 30.70 & \textbf{100.0} & \textbf{86.47} \\
relief 2      
& 11.8 & 11.8 & 50.26 & 47.35 & 69.9 & 70.3 & 40.9 & 39.1 & 57.4 & 76.1 & \textbf{100.0} & 75.7 & \textbf{100.0} & \textbf{95.9} & 45.47 & 8.64 & \textbf{100.0} & \textbf{73.47} \\
terrace       
& \textbf{100.0} & \textbf{100.0} & 83.74 & 79.02 & 38.7 & 29.6 & \textbf{100.0} & 99.6 & \textbf{100.0} & \textbf{100.0} & \textbf{100.0} & 97.2 & \textbf{100.0} & 95.6 & 90.17 & 61.44 & \textbf{100.0} & \textbf{92.82} \\
terrains      
& \textbf{100.0} & 99.5 & 55.90 & 50.17 & 70.4 & 54.9 & \textbf{100.0} & 91.9 & 58.2 & 52.5 & 91.6 & 53.8 & \textbf{100.0} & \textbf{97.3} & 37.64 & 9.35 & \textbf{100.0} & \textbf{90.53} \\
\midrule
Average       
& 49.0 & 47.8 & 58.40 & 47.70 & 65.4 & 58.9 & 74.2 & 70.7 & 81.2 & 79.7 & 98.2 & 81.9 & \textbf{100.0} & \textbf{91.3} & 77.69 & 35.55 & \textbf{91.58} & \textbf{73.27} \\
\bottomrule
\end{tabular}
}
\label{tab:eth3d}
\end{table}
\vspace{-2mm}

%% file: tables/abla.tex
\begin{table*}[t]
  \centering
  \caption{Ablation study of different components in our method.}
  \label{tab:pose_comparison}
  \resizebox{\textwidth}{!}{
    \begin{tabular}{lccc|ccc}
      \toprule
      \textbf{Method} & \textbf{Init} & \textbf{Motion Avg.} & \textbf{Ours} & \textbf{Coarse Track.} & \textbf{VGGT-Track. \citep{wang2025vggt}} & \textbf{RomaV2-Track.~\cite{edstedt2025roma}} \\
      \midrule
      RRA@1 ($\uparrow$)   & 69.70 & 70.20 & \textbf{90.80} & 67.11 & 59.29 & 57.12 \\
      RTA@1 ($\uparrow$)   & 33.64 & 34.75 & \textbf{73.27} & 48.56 & 29.35 & 44.78 \\
      Speed (FPS) ($\uparrow$) & \textbf{3.34} & 2.77 & 2.06 & 2.54 & 1.14 & 0.82 \\
      \bottomrule
    \end{tabular}
  }
  \label{tab:abla}
  \vspace{-2mm}
\end{table*}

%% file: sections/05_conclusion.tex
\section{Conclusion}
\label{sec:conclu}

We present a foundation-model-guided global SfM framework that transforms feed-forward 3D predictions into optimizable geometric constraints. Our method leverages the predicted geometry and cross-view features of a frozen foundation model to build reliable correspondences and an initial pose graph. With a keyframe-based sliding window strategy, these constraints are propagated across long sequences and unordered image collections. Global pose estimation and bundle adjustment then refine the poses, reduce scale inconsistencies, and recover dense geometry. Experiments on diverse benchmarks demonstrate that our framework combines the robustness of 3D foundation models with global SfM-style optimization, achieving efficient and high-fidelity reconstruction.

\noindent
\textbf{Future Work.}
Our current sliding-window strategy adopts a fixed window size. Adapting the window size according to motion type, visual overlap, and matching confidence may further improve efficiency and robustness. In addition, the matching head still depends on the geometric predictions of the underlying foundation model. 
Reducing this dependence and improving robustness to imperfect initial geometry remain important directions for future research.

%% file: sections/XX_supp.tex
\appendix

In this appendix, we provide additional details and results to complement the main paper. Sec.~\ref{sec:sup:net} describes the network architecture, including the Pi3X backbone, matching network, and refinement module. Sec.~\ref{sec:sup:loss} presents the loss functions and training protocol. Sec.~\ref{sec:sup:exp} provides further experimental details, including inference settings, baseline configurations, additional quantitative and qualitative results, and failure case analysis.

\section{Network Architecture Details}
\label{sec:sup:net}
\noindent
\textbf{Pi3X Backbone.}
Given an input image set $\mathcal{I}=\{I_i\}_{i=1}^{N}$ with resolution $H\times W$, our network takes images of shape $N\times H\times W\times 3$ as input. We first feed the images into the frozen Pi3X~\cite{wang2025pi3} backbone. Following Pi3X, each image is encoded by a DINOv2 encoder, producing encoder tokens
\begin{equation}
    \mathbf{E} \in \mathbb{R}^{N\times H'\times W'\times C},
    \quad H'=\frac{H}{14}, \quad W'=\frac{W}{14}.
\end{equation}
The encoder processed by alternating frame-wise and global-wise attention blocks~\cite{wang2025vggt}. The decoder contains 18 attention blocks in total. The outputs of the last frame-wise and global-wise attention layers are concatenated, yielding the final geometry token
\begin{equation}
    \mathbf{H} \in \mathbb{R}^{N\times H'\times W'\times 2C}.
\end{equation}
The geometry token is further decoded by the original Pi3X prediction heads, including the point decoder, camera decoder, and metric decoder, to predict local point maps, camera poses, and metric scales, respectively. We refer readers to Pi3X~\cite{wang2025pi3} for more details of the backbone architecture.

\noindent
\textbf{Matching Decoder.}
Our coarse dense matching branch operates on the geometry token $\mathbf{H}$ produced by the Pi3X decoder. We introduce a lightweight matching decoder, denoted as $\mathrm{Dec}_{\mathrm{match}}$, which consists of 5 attention layers. It transforms the $2C$-dimensional geometry token into a compact matching token:
\begin{equation}
    \mathbf{Z}
    =
    \mathrm{Dec}_{\mathrm{match}}(\mathbf{H}),
    \quad
    \mathbf{Z}\in\mathbb{R}^{N\times H'\times W'\times C}.
\end{equation}
In our experiments, we find this matching decoder to be important. Directly reducing the channel dimension with a linear projection leads to significantly worse matching quality, suggesting that additional attention layers are needed to adapt the frozen geometry representation to dense correspondence prediction.

During training, we use the first image as the reference image and predict dense warps from it to all other images. During inference, thanks to the permutation-equivariant property of Pi3X, any image can be selected as the reference. Multiple reference images can also be handled by expanding the batch dimension.

\noindent
\textbf{Multi-view Match Embedding.}
We next construct multi-view match embeddings for dense warping. Let $ M = H'W',$
and flatten the spatial dimensions of the matching tokens. During training, we use the image $I_a$ as the reference image. Its matching token is denoted as
\begin{equation}
    \mathbf{Z}^{a}=\{\mathbf{z}^{a}_{n}\}_{n=1}^{M},
    \quad
    \mathbf{z}^{a}_{n}\in\mathbb{R}^{C}.
\end{equation}
For each target image $I_b$, $b\in\{1,\cdots,N\}\setminus\{a\}$, we denote its matching token as
\begin{equation}
    \mathbf{Z}^{b}=\{\mathbf{z}^{b}_{m}\}_{m=1}^{M},
    \quad
    \mathbf{z}^{b}_{m}\in\mathbb{R}^{C}.
\end{equation}

For each target-reference pair $(a,b)$, we compute a patch-level similarity matrix from target patches to reference patches:
\begin{equation}
    \mathbf{S}^{a\rightarrow b}_{mn}
    =
    \exp
    \left(
    \frac{1}{\tau}
    \mathrm{cosim}
    \left(
    \mathbf{z}^{a}_{m},
    \mathbf{z}^{b}_{n}
    \right)
    \right),
    \quad
    \mathbf{S}^{a\rightarrow b}\in\mathbb{R}^{M\times M},
\end{equation}
where $\tau=1/10$ is the temperature following RoMa, and $\mathrm{cosim}(\mathbf{x},\mathbf{y})$ denotes cosine similarity:
\begin{equation}
    \mathrm{cosim}(\mathbf{x},\mathbf{y})
    =
    \frac{\mathbf{x}^{\top}\mathbf{y}}
    {\|\mathbf{x}\|\|\mathbf{y}\|}.
\end{equation}
Computing this similarity for all target images yields
\begin{equation}
    \mathbf{S}^{a\rightarrow\mathcal{B}}
    =
    \left\{
    \mathbf{S}^{a\rightarrow b}
    \right\}_{b\in\mathcal{B}}
    \in
    \mathbb{R}^{(N-1)\times M\times M},
    \quad
    \mathcal{B}=\{1,\cdots,N\}\setminus\{a\}.
\end{equation}

Following RoMa, we convert the similarity matrix into match embeddings using Fourier positional embeddings of the reference image coordinates. For each reference patch coordinate $\mathbf{x}^{a}_{n}\in\mathbb{R}^{2}$, we define
\begin{equation}
    \boldsymbol{\chi}^{a}_{n}
    =
    \cos(2\pi \omega \mathbf{W}\mathbf{x}^{a}_{n})
    \oplus
    \sin(2\pi \omega \mathbf{W}\mathbf{x}^{a}_{n})
    \in \mathbb{R}^{C},
\end{equation}
where $\oplus$ denotes concatenation, $\omega=1$, and $\mathbf{W}$ is a fixed non-learnable Gaussian matrix with a compatible output dimension. The match embedding for the target patch $m$ in the image $I_b$ is then calculated by aggregating the Fourier embeddings of the reference patches according to the similarity scores:
\begin{equation}
    \boldsymbol{\chi}^{a\rightarrow b}_{m}
    =
    \sum_{n=1}^{M}
    \mathbf{S}^{a\rightarrow b}_{mn}
    \boldsymbol{\chi}^{a}_{n},
    \quad
    \boldsymbol{\chi}^{a\rightarrow b}_{m}\in\mathbb{R}^{C}.
\end{equation}
Therefore, the multi-view match embeddings for all target images are
\begin{equation}
    \boldsymbol{\chi}^{a\rightarrow\mathcal{B}}
    =
    \left\{
    \boldsymbol{\chi}^{a\rightarrow b}
    \right\}_{b\in\mathcal{B}}
    \in
    \mathbb{R}^{(N-1)\times H'\times W'\times C}.
\end{equation}
This formulation extends the two-view match embedding of RoMa to the multi-view setting by computing reference-to-target dense correspondence embeddings for all images in the local window.

\noindent
\textbf{DPT Matching Head.}
Given the reference-to-target match embeddings $\boldsymbol{\chi}^{ a\rightarrow\mathcal{B}}$, we build pair-wise matching features for each target image $I_b$ by combining the target matching token with its corresponding match embedding:
\begin{equation}
    \mathbf{F}^{a\rightarrow b}
    =
    \mathrm{Proj}
    \left(
    \mathbf{Z}^{b}
    \oplus
    \boldsymbol{\chi}^{a\rightarrow b}
    \right),
    \quad
    \mathbf{F}^{a\rightarrow b}
    \in
    \mathbb{R}^{H'\times W'\times C},
\end{equation}
where $\oplus$ denotes channel-wise concatenation and $\mathrm{Proj}(\cdot)$ is a linear projection. Intuitively, $\mathbf{Z}^{b}$ provides the local image representation of the target frame, while $\boldsymbol{\chi}^{a\rightarrow b}$ encodes the soft correspondence distribution from the reference image coordinates patch to each target.

For all target images in the local window, the pair-wise features are stacked as
\begin{equation}
    \mathbf{F}^{a\rightarrow \mathcal{B}}
    =
    \left\{
    \mathbf{F}^{a\rightarrow b}
    \right\}_{b\in\mathcal{B}}
    \in
    \mathbb{R}^{(N-1)\times H'\times W'\times C}.
\end{equation}
We feed these features, together with the multi-scale encoder features from the Pi3X backbone, into a Dense Prediction Transformer head:
\begin{equation}
    \left(
    \mathbf{W}^{a\rightarrow\mathcal{B}},
    \mathbf{p}^{a\rightarrow\mathcal{B}}
    \right)
    =
    \mathrm{DPT}_{\mathrm{match}}
    \left(
    \mathbf{F}^{a\rightarrow\mathcal{B}},
    \mathbf{E}
    \right).
\end{equation}
The output is a dense warp from each target image to the reference image and its confidence:
\begin{equation}
    \mathbf{W}^{a\rightarrow\mathcal{B}}
    \in
    \mathbb{R}^{(N-1)\times 2\times \frac{H}{4}\times \frac{W}{4}},
    \quad
    \mathbf{p}^{a\rightarrow\mathcal{B}}
    \in
    \mathbb{R}^{(N-1)\times 1\times \frac{H}{4}\times \frac{W}{4}}.
\end{equation}

We set the finest resolution of the DPT head to one quarter of the input image resolution. Following the standard DPT design, we use a scratch dimension of 256 and output dimensions $[256,512,1024,1024]$ for feature strides $[4,8,16,32]$, respectively. The final coarse warp and confidence are predicted at stride 4, and are then passed to the refinement module to recover full-resolution correspondences.

\noindent
\textbf{Refinement Module.}
The coarse matching head predicts warps at stride 4. To recover full-resolution correspondences, we further employ a lightweight refinement module following the coarse-to-fine design of RoMaV2~\cite{edstedt2025roma}. Since our coarse prediction is already produced at stride 4, we only keep refinement stages at strides $\{4,2,1\}$, which avoids unnecessary high-level refinement and reduces both feature extraction and inference cost.

We extract fine image features at three resolutions:
\begin{equation}
    \boldsymbol{\phi}_4 \in \mathbb{R}^{\frac{H}{4}\times\frac{W}{4}\times192},
    \quad
    \boldsymbol{\phi}_2 \in \mathbb{R}^{\frac{H}{2}\times\frac{W}{2}\times48},
    \quad
    \boldsymbol{\phi}_1 \in \mathbb{R}^{H\times W\times12}.
\end{equation}
These features are linearly projected before being passed to the corresponding refinement blocks. At each stride $s\in\{4,2,1\}$, the refiner takes the current warp estimate $\mathbf{W}^{a\rightarrow b}$ and predicts a residual update. Its input is constructed by concatenating the fine features from the reference and target images, a positional displacement embedding, and a local correlation feature around the current matched location:
\begin{equation}
    \mathbf{F}_{s}^{a\rightarrow b}
    =
    \boldsymbol{\phi}_{s}^{a}
    \oplus
    \boldsymbol{\phi}_{s}^{b}(\mathbf{W}^{a\rightarrow b})
    \oplus
    g_s\!\left(\mathbf{W}^{a\rightarrow b}-\mathbf{x}^{a}\right)
    \oplus
    \mathrm{Corr}_{s}
    \left(
    \boldsymbol{\phi}_{s}^{a},
    \boldsymbol{\phi}_{s}^{b},
    \mathbf{W}^{a\rightarrow b}
    \right),
\end{equation}
where $\oplus$ denotes channel-wise concatenation, $\boldsymbol{\phi}_{s}^{b}(\mathbf{W}^{a\rightarrow b})$ denotes target features sampled at the warped coordinates, $\mathbf{x}^{a}$ is the reference pixel coordinate, and $g_s(\cdot)$ is a linear projection of the current displacement. The local correlation term computes a small neighborhood correlation in the target feature map centered at the current warp location, providing fine-grained matching evidence for residual correction.

Each refinement block predicts a residual warp and confidence update:
\begin{equation}
    (\Delta \mathbf{W}_{s}^{a\rightarrow b}, \Delta p_{s}^{a\rightarrow b})
    =
    \mathrm{Refine}_{s}
    \left(
    \mathbf{F}_{s}^{a\rightarrow b}
    \right),
\end{equation}
and the warp is progressively updated from stride 4 to stride 1:
\begin{equation}
    \mathbf{W}_{s}^{a\rightarrow b}
    \leftarrow
    \mathrm{upsample}
    \left(
    \mathbf{W}_{2s}^{a\rightarrow b}
    \right)
    +
    \Delta \mathbf{W}_{s}^{a\rightarrow b}.
\end{equation}
For local correlation, we use window sizes $[k_4,k_2,k_1]=[7,3,0]$, where $k_1=0$ means that no local correlation is used at full resolution. The internal structure of each refinement block follows a compact convolutional design with depthwise convolution, normalization, non-linearity, and pointwise projection. This refinement stage substantially improves the localization accuracy of the dense warp while keeping the additional computation moderate.

\section{Loss Function and Training Details}
\label{sec:sup:loss}
We train the matching branch using three losses: an auxiliary patch-level negative log-likelihood loss on the coarse similarity matrix, a dense warp regression loss, and a confidence supervision loss. During training, we use the first image $I_a$ in each local window as the reference image, and treat the remaining images $\{I_b\}_{b\in\mathcal{B}}$ as target images, where $\mathcal{B}=\{1,\ldots,N\}\setminus\{a\}$.

\noindent
\textbf{Ground-truth Warp and Validity.}
For each reference-target pair $(a,b)$, we generate the ground-truth warp from the reference image $I_a$ to the target image $I_b$ using the ground-truth depth and camera poses. For a reference pixel $\mathbf{x}^{a}$ with depth $z^{a}(\mathbf{x}^{a})$, its corresponding point is first back-projected to the camera coordinate system of $I_a$ and then transformed to the target view:
\begin{equation}
    \tilde{\mathbf{x}}^{a\rightarrow b}
    =
    \mathbf{K}^{b}
    \left(
    \mathbf{R}^{a\rightarrow b}
    \left(\mathbf{K}^{a}\right)^{-1}
    \tilde{\mathbf{x}}^{a}
    z^{a}(\mathbf{x}^{a})
    +
    \mathbf{t}^{a\rightarrow b}
    \right),
\end{equation}
where $\tilde{\mathbf{x}}^{a}$ denotes the homogeneous pixel coordinate, and $(\mathbf{R}^{a\rightarrow b},\mathbf{t}^{a\rightarrow b})$ is the relative pose from $I_a$ to $I_b$. The projected pixel and its corresponding depth in $I_b$ are
\begin{equation}
    \mathbf{x}^{a\rightarrow b}
    =
    \pi\left(\tilde{\mathbf{x}}^{a\rightarrow b}\right),
    \quad
    z^{a\rightarrow b}
    =
    \left[
    \mathbf{R}^{a\rightarrow b}
    \left(\mathbf{K}^{a}\right)^{-1}
    \tilde{\mathbf{x}}^{a}
    z^{a}(\mathbf{x}^{a})
    +
    \mathbf{t}^{a\rightarrow b}
    \right]_3,
\end{equation}
where $\pi(\cdot)$ denotes the projection from homogeneous coordinates to image coordinates, and $[\cdot]_3$ extracts the depth component. The ground-truth warp is therefore defined as
\begin{equation}
    \mathbf{W}^{*\,a\rightarrow b}(\mathbf{x}^{a})
    =
    \mathbf{x}^{a\rightarrow b}.
\end{equation}

We define the confidence label according to target-view depth consistency. A correspondence is assigned positive confidence if the projected pixel lies inside the target image, both the projected depth and the sampled target depth are positive, and their relative depth difference is smaller than a threshold:
\begin{equation} y^{a\rightarrow b}(\mathbf{x}^{a}) = \mathbf{1} \left[ \mathbf{x}^{a\rightarrow b}\in\mathcal{D}^{b}, \quad z^{a\rightarrow b}>0, \quad \frac{ \left| z^{b}\!\left(\mathbf{x}^{a\rightarrow b}\right) - z^{a\rightarrow b} \right| }{ z^{b}\!\left(\mathbf{x}^{a\rightarrow b}\right) } < \tau_z \right], \end{equation}
where $\mathcal{D}^{b}$ is the image domain of $I_b$, $z^{b}(\mathbf{x}^{a\rightarrow b})$ is the target-view depth sampled at the projected location, and $\tau_z=0.05$ in our training.

\begin{figure}
    \centering
    \includegraphics[width=1.00\linewidth]{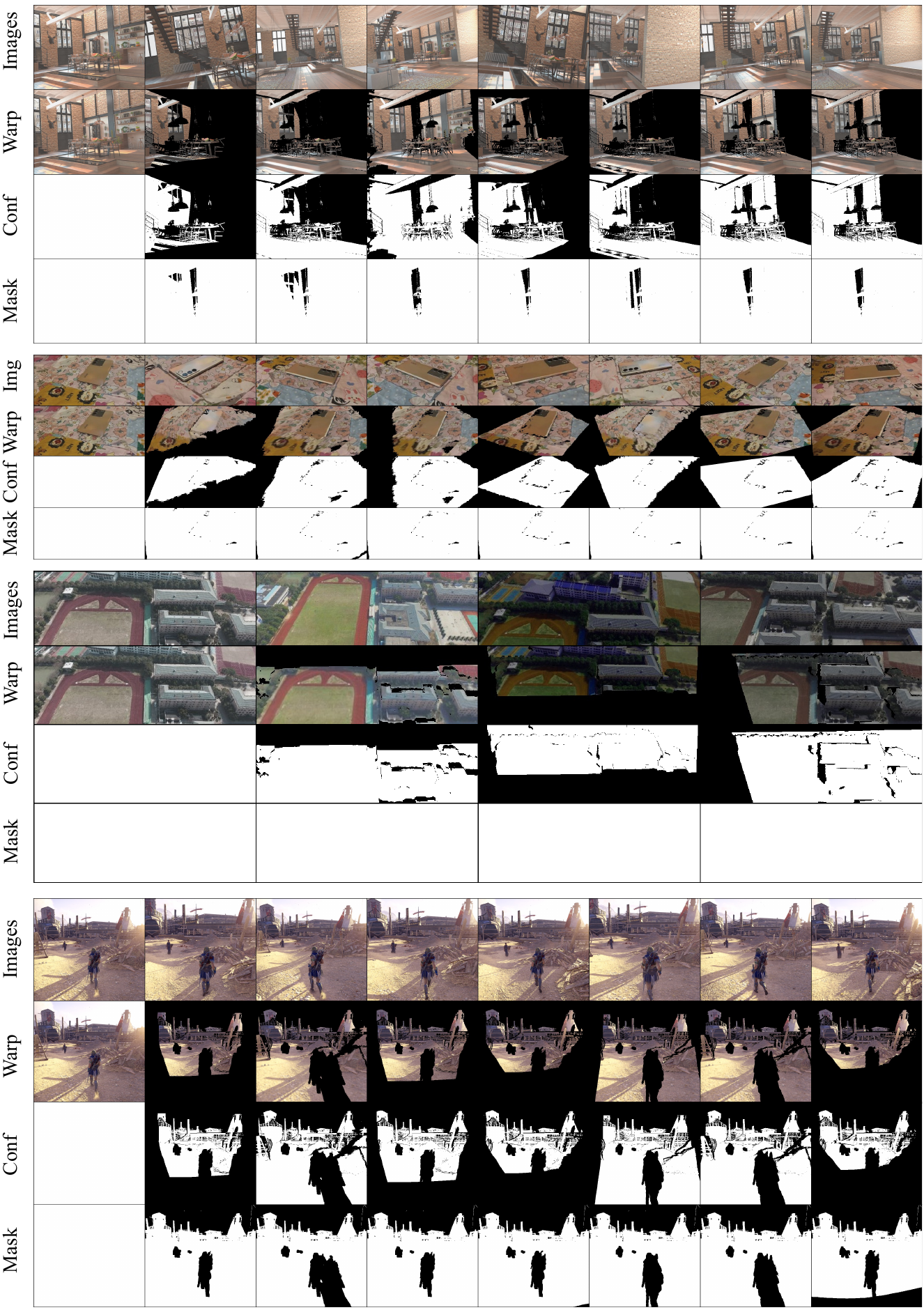}
    \caption{Qualitative results on various datasets. We visualize the original images, warped results, confidence maps, and masks. The rows from top to bottom correspond to the Hypersim~\cite{roberts2021hypersim}, WildRGBD~\cite{xia2024rgbd}, BlendedMVS~\cite{yao2020blendedmvs}, and OmniWorld~\cite{zhou2025omniworld} datasets, respectively.}
    \label{fig:training_data}
    \vspace{-2mm}
\end{figure}

We further define a training mask to specify where the confidence loss and warp loss is applied. The mask is valid when the reference depth is positive and either the projected point falls inside the target image with positive projected and sampled depths, or the projected point falls outside the target image. The latter case is included to explicitly supervise out-of-bound correspondences with zero confidence. Formally,
\begin{equation}
\begin{aligned}
    m^{a\rightarrow b}(\mathbf{x}^{a})
    =
    \mathbf{1}
    \Big[
    & z^{a}(\mathbf{x}^{a})>0
    \quad \mathrm{and} \quad
    \Big((
    \mathbf{x}^{a\rightarrow b}\in\mathcal{D}^{b}
    \quad \mathrm{and} \quad
    z^{a\rightarrow b}>0
    \quad \mathrm{and} \quad
    z^{b}\!\left(\mathbf{x}^{a\rightarrow b}\right)>0)
    \\
    & \quad \mathrm{or} \quad
    \mathbf{x}^{a\rightarrow b}\notin\mathcal{D}^{b}
    \Big)
    \Big].
\end{aligned}
\end{equation}
The warp regression loss is evaluated only on pixels with positive confidence label and positive mask label, while the confidence loss is evaluated on pixels selected by this mask. We visualize the original images, warped results, masks, and confidence maps in Fig.~\ref{fig:training_data}.

\noindent
\textbf{Auxiliary NLL Loss.}
To stabilize training of the coarse matching head, we add an auxiliary negative log-likelihood loss on the similarity matrix. Recall that for each target-reference pair $(b,a)$, the coarse matching decoder produces a similarity matrix $\mathbf{S}^{a\rightarrow b}\in\mathbb{R}^{M\times M}$, where $M=H'W'$. For each target patch $m$, we determine the nearest reference patch index $n_m^*$ according to the ground-truth warp. We then apply a row-wise Softmax to $\mathbf{S}^{a\rightarrow b}$ and minimize
\begin{equation}
    \mathcal{L}_{\mathrm{NLL}}^{a\rightarrow b}
    =
    -\frac{1}{|\Omega_{\mathrm{patch}}^{a\rightarrow b}|}
    \sum_{m\in\Omega_{\mathrm{patch}}^{a\rightarrow b}}
    \log
    \left(
    \mathrm{Softmax}
    \left(
    \mathbf{S}^{a\rightarrow b}_{m:}
    \right)_{n_m^*}
    \right),
\end{equation}
where $\Omega_{\mathrm{patch}}^{a\rightarrow b}$ denotes the set of valid target patches. This loss encourages the coarse similarity matrix to assign high probability to the correct reference patch before dense refinement.

\noindent
\textbf{Warp Loss.}
Given the predicted dense warp $\hat{\mathbf{W}}^{a\rightarrow b}$, we supervise it with a generalized Charbonnier loss. Let
\begin{equation}
    \mathbf{r}^{a\rightarrow b}(\mathbf{x})
    =
    \hat{\mathbf{W}}^{a\rightarrow b}(\mathbf{x})
    -
    \mathbf{W}^{*\,a\rightarrow b}(\mathbf{x})
\end{equation}
denote the warp residual at pixel $\mathbf{x}$. The warp loss is defined as
\begin{equation}
    \mathcal{L}_{\mathrm{warp}}^{a\rightarrow b}
    =
    \frac{1}{|\Omega^{a\rightarrow b}|}
    \sum_{\mathbf{x}\in\Omega^{a\rightarrow b}}
    \left(
    \|\mathbf{r}^{a\rightarrow b}(\mathbf{x})\|_2^2+\epsilon^2
    \right)^{\alpha/2},
\end{equation}
where $\Omega^{a\rightarrow b}=\{\mathbf{x}\mid y^{a\rightarrow b}(\mathbf{x})=1\}$ is the set of valid pixels, $\epsilon$ is a small constant, and $\alpha$ controls the robustness of the penalty. We apply this loss to the coarse prediction as well as to all refinement stages.

\noindent
\textbf{Confidence Loss.}
In addition to the warp, the network predicts a confidence map $\hat{p}^{a\rightarrow b}\in[0,1]$. We supervise it using the validity mask defined above. Specifically, we use a binary cross-entropy loss
\begin{equation}
    \mathcal{L}_{\mathrm{conf}}^{a\rightarrow b}
    =
    -\frac{1}{|\mathcal{D}|}
    \sum_{\mathbf{x}\in\mathcal{D}}
    \left[
    y^{a\rightarrow b}(\mathbf{x})
    \log \hat{p}^{a\rightarrow b}(\mathbf{x})
    +
    \left(1-y^{a\rightarrow b}(\mathbf{x})\right)
    \log \left(1-\hat{p}^{a\rightarrow b}(\mathbf{x})\right)
    \right].
\end{equation}
This loss encourages the predicted confidence to reflect whether a correspondence is geometrically valid and depth-consistent.

\noindent
\textbf{Total Loss.}
The final training objective sums the three terms over all target images in the local window:
\begin{equation}
    \mathcal{L}
    =
    \sum_{b\in\mathcal{B}}
    \lambda_{\mathrm{NLL}}
    \mathcal{L}_{\mathrm{NLL}}^{a\rightarrow b}
    +
    \lambda_{\mathrm{warp}}
    \mathcal{L}_{\mathrm{warp}}^{a\rightarrow b}
    +
    \lambda_{\mathrm{conf}}
    \mathcal{L}_{\mathrm{conf}}^{a\rightarrow b},
\end{equation}
where $\lambda_{\mathrm{NLL}}$, $\lambda_{\mathrm{warp}}$, and $\lambda_{\mathrm{conf}}$ balance the three losses.

\noindent
\textbf{Training Details.}
To preserve the geometric prior of the foundation model, we freeze the DINOv2 encoder and the original Pi3X geometry heads, and train only the matching decoder, the DPT matching head, and the refinement modules. We train the model on 16 NVIDIA H20 GPUs with gradient accumulation of 2 steps, resulting in an effective batch size equivalent to 32 GPUs. The coarse matching stage and the refinement stage are trained separately, each for 32K iterations. The overall training takes about 4--5 days.

For the coarse stage, we randomly sample 2 to 48 frames from each training sequence. The input frames are resized such that the longer image side is at most 518 pixels, and the aspect ratio is randomly sampled between 0.33 and 1.0. We apply standard data augmentation, including color jittering, Gaussian blur, and random grayscale conversion. The model is optimized with a cosine learning-rate schedule, using a peak learning rate of $1\times10^{-4}$ and 2K warm-up iterations.

The refinement stage follows a similar training protocol, except that we sample 2 to 24 frames from each sequence. We initialize the refinement module from the RoMa v2 refinement checkpoint and train it with a cosine learning-rate schedule using a peak learning rate of $5\times10^{-5}$ and 2K warm-up iterations.

\noindent
\textbf{Training Data.}
We train the matching branch on a diverse mixture of synthetic and real-world multi-view datasets, including Aria Synthetic Environments, Aria Digital Twin~\cite{pan2023aria}, DL3DV~\cite{ling2024dl3dv}, Co3Dv2~\cite{reizenstein21co3d}, ARKitScenes~\cite{baruch2021arkitscenes}, BlendMVS~\cite{yao2020blendedmvs}, HyperSim~\cite{roberts2021hypersim}, MegaDepth~\cite{li2018megadepth}, MVS-Synth~\cite{DeepMVS}, ScanNet~\cite{dai2017scannet}, ScanNet++~\cite{yeshwanth2023scannet++}, TartanAir~\cite{wang2020tartanair}, Virtual KITTI~\cite{cabon2020virtual}, Unreal4K~\cite{Tosi2021CVPR}, WildRGB-D~\cite{xia2024rgbd}, MatrixCity~\cite{li2023matrixcity}, and OmniWorld~\cite{zhou2025omniworld}. These datasets cover indoor and outdoor scenes, object-centric captures, synthetic environments, driving scenarios, and large-scale urban scenes, providing diverse camera motions, appearance variations, and geometric layouts for robust correspondence learning.

\section{Experiments}
\label{sec:sup:exp}
\input{tables/tnt_full}

\begin{figure}
    \centering
    \includegraphics[width=1.00\linewidth]{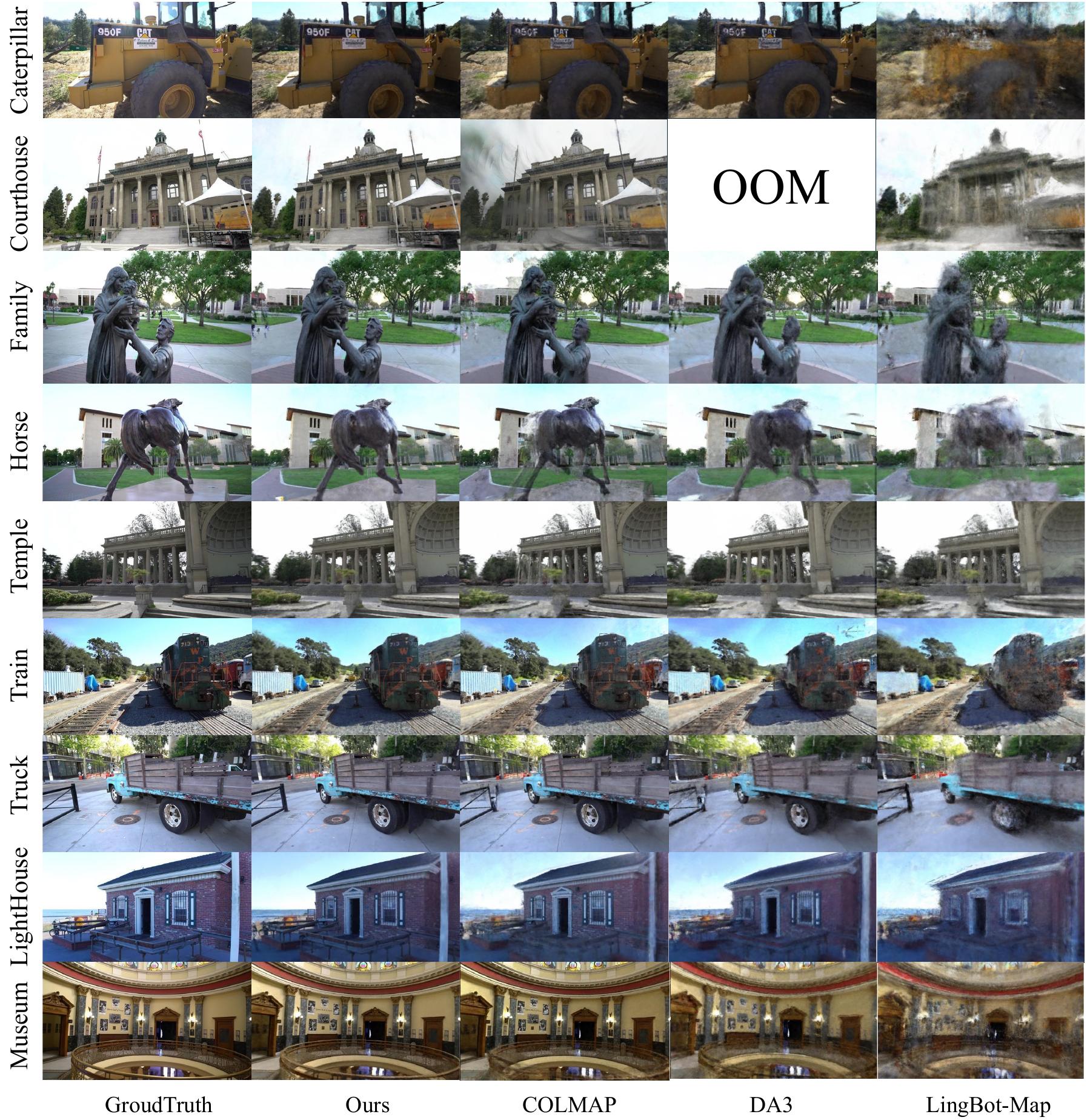}
    \caption{Additional qualitative comparisons on Tanks and Temples. Our method produces sharper and more faithful rendered views with fewer pose-induced artifacts.}
    \label{fig:tntpsnr2}
    \vspace{-2mm}
\end{figure}

\begin{figure}
    \centering
    \includegraphics[width=1.00\linewidth]{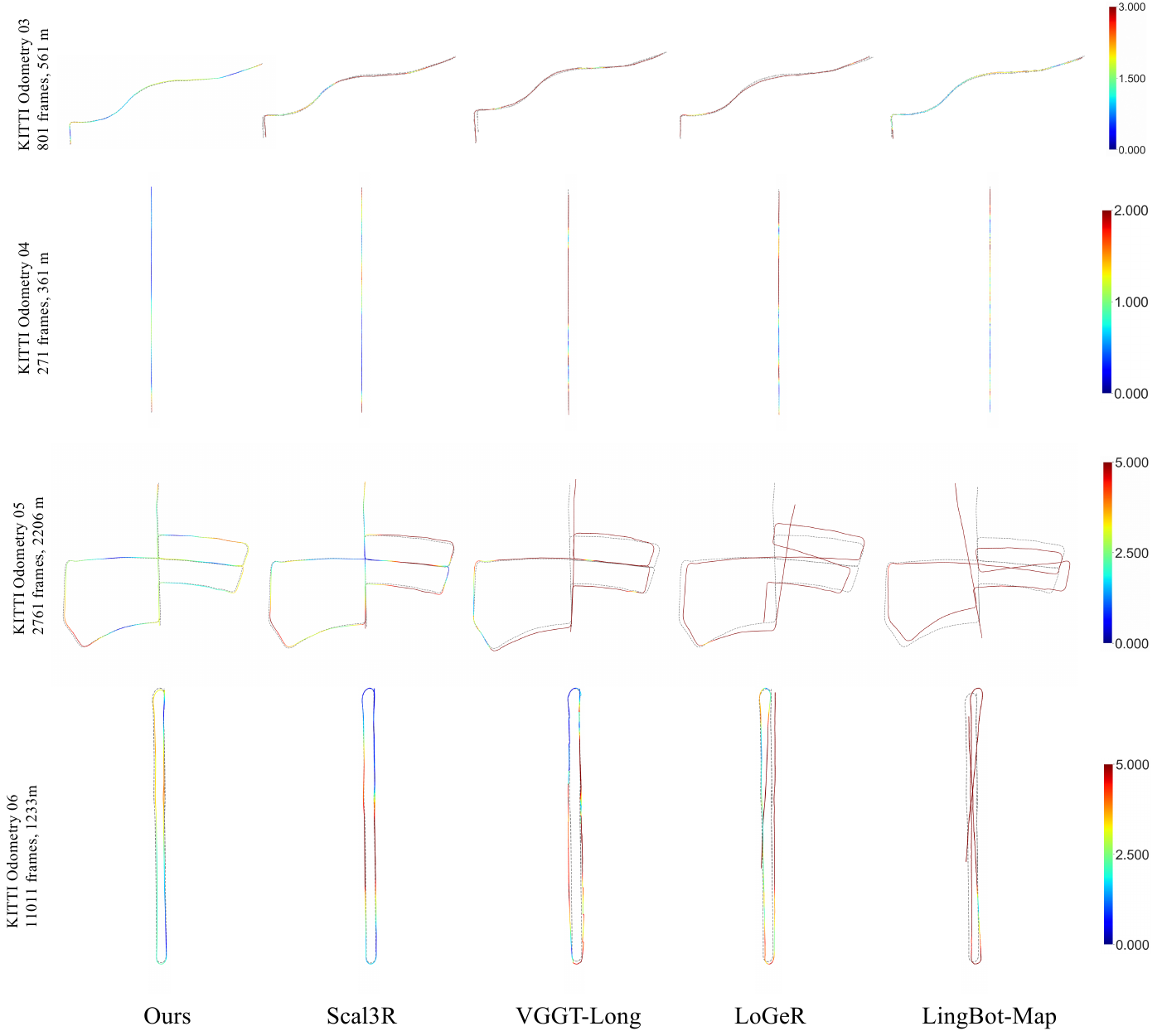}
    \caption{Additional KITTI trajectory visualizations. Our method better preserves global trajectory shape under long forward-driving motion.}
    \label{fig:kitti03040506}
    \vspace{-2mm}
\end{figure}

\begin{figure}
    \centering
    \includegraphics[width=1.00\linewidth]{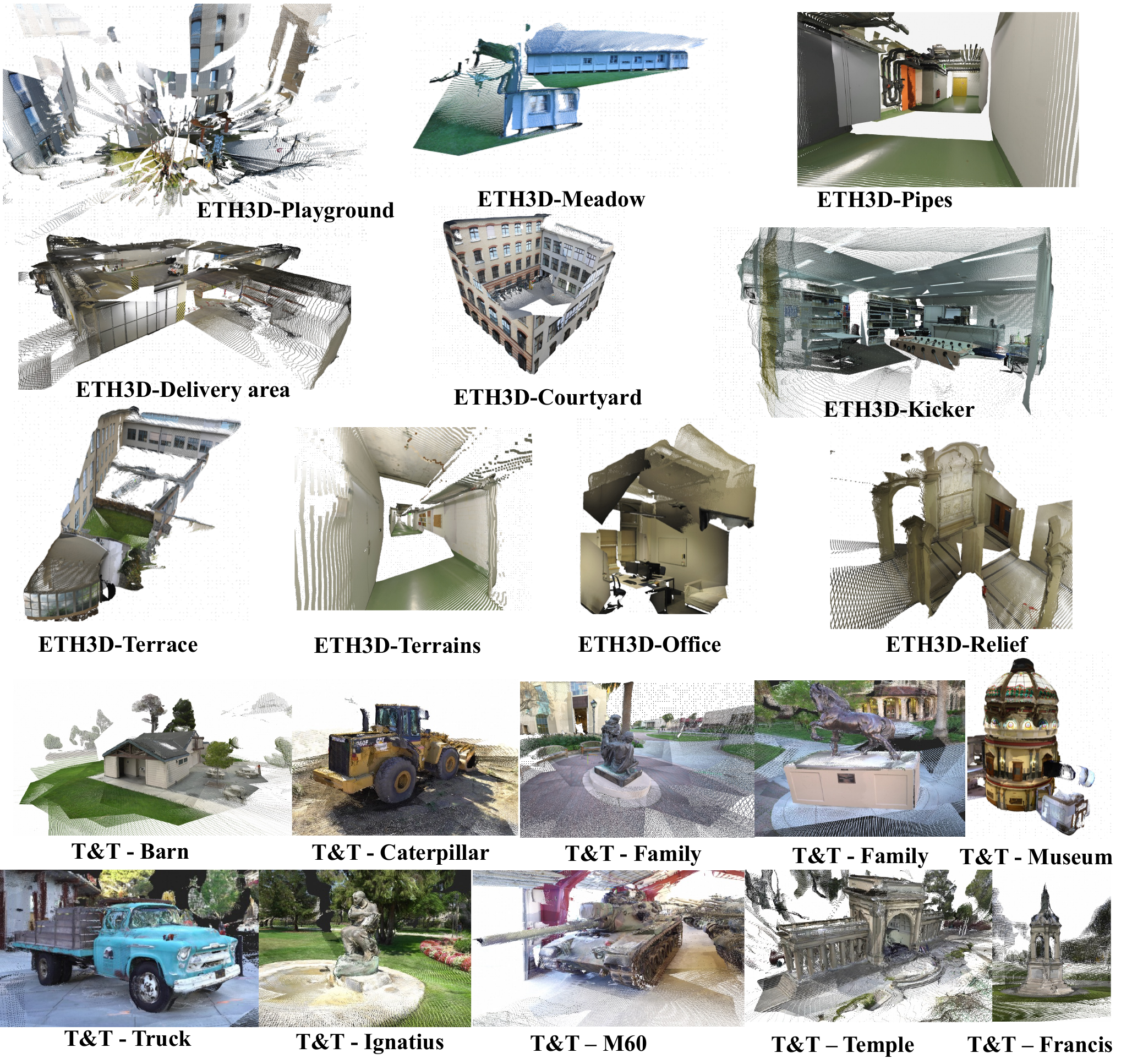}
    \caption{Dense point cloud reconstructions produced by our optimized poses and recovered dense depths.}
    \label{fig:densepoints}
    \vspace{-2mm}
\end{figure}
\subsection{Inference Details}

Most experiments are conducted on a single NVIDIA L20 GPU with 48GB memory. Unless otherwise specified, we use a sliding window size of $N=20$ and shift the window by 10 frames. For each keyframe, we sample 512 tracking points by default. For large-scale KITTI sequences, we reduce the number of sampled points to 256 to lower memory consumption. We use a warp confidence threshold of 0.6 and a depth confidence threshold of 0.1 for filtering reliable correspondences.

\noindent
\textbf{Sliding-window association.}
During inference, each window is processed as described in the main paper. We select keyframes according to the valid reprojection ratio. Specifically, if the ratio of valid projected pixels from the current frame to all existing keyframes in the window is below 0.2, the current frame is selected as a new keyframe. In other words, a frame is promoted to a keyframe when fewer than 20\% of its pixels can be reliably projected to the current keyframe set. Due to the half-window overlap, most keyframes participate in two neighboring windows. Except for keyframes near the beginning and the end of the sequence, each keyframe can be associated with approximately $20+20-10=30$ frames. For frames appearing in the overlap, we use the matching results from their first inference to avoid duplicate tracks.

\noindent
\textbf{Track sampling.}
After obtaining dense image warps, we sample tracking points from each keyframe. We first filter pixels using the depth confidence threshold, and then randomly sample the required number of pixels from the remaining reliable regions. These points are propagated to other frames using the predicted dense warp, and matches with warp confidence lower than 0.6 are discarded. This produces sparse but reliable tracks for subsequent pose initialization and optimization.

\noindent
\textbf{Retrieval-based loop association.}
In addition to local sliding-window associations, we perform image retrieval-based loop candidate discovery over all keyframes. We extract SALAD~\cite{Izquierdo_CVPR_2024_SALAD} image features for each keyframe and compute pairwise cosine similarities. Two keyframes are considered to have potential visual overlap if their similarity is larger than 0.5. For each keyframe, we collect all retrieved keyframes whose similarity exceeds this threshold to form an additional matching window, and use the first image in this window as the tracking reference. The resulting long-range matches are added to the optimization graph as loop constraints. This retrieval step is not only useful for explicit loop closure; it can also connect temporally distant but spatially close keyframes, especially in slow-motion regions where adjacent sliding windows may have limited viewpoint change.

\noindent
\textbf{Optimization settings.}
In the optimization stage, we support both shared and per-frame camera intrinsics. By default, we assume that all frames in a sequence share the same intrinsic parameters, while per-frame intrinsics can be enabled when necessary. For large-scale KITTI sequences, after retrieval-based loop detection, we first run pose graph optimization using relative poses from loop keyframe pairs and adjacent-frame constraints from the initialization. The optimized poses are then used to initialize the final global refinement.

\noindent
\textbf{Window size adjustment.}
As discussed in Sec.~\ref{sec:conclu}, our current implementation uses a fixed window size by default, and the matching quality can be affected by the geometric predictions of the underlying foundation model. In most scenes, the default setting of $N=20$ works well. However, for a few challenging sequences, small local windows may provide insufficient context for Pi3X, leading to ambiguous geometric predictions such as inaccurate scale or unstable point maps. In such cases, using a larger window provides more multi-view context and improves the stability of the initial geometry. Therefore, we slightly adjust the window size for several difficult sequences. Specifically, we use $N=120$ for the T\&T Auditorium and Courtroom scenes, and $N=200$ for KITTI sequence 02. This adjustment is only applied to ensure stable inference on these challenging cases, and does not change the overall pipeline or optimization procedure.

\noindent
\textbf{GPU-based optimization.}
The global positioning and bundle adjustment stages can become a computational bottleneck if implemented with conventional CPU-based solvers, especially for long sequences with many tracks. Following recent GPU-accelerated SfM and bundle adjustment systems~\cite{zhong2026instantsfm,zhan2026bundle}, we implement the optimization stage in a PyTorch-compatible GPU framework. In particular, the residuals and Jacobians for both global positioning and BA are constructed on GPU, and the sparse normal-equation structure is solved with batched parallel operations. This GPU-based implementation allows our optimization module to better match the efficiency of the feed-forward foundation model inference while preserving the accuracy of global geometric refinement.

\subsection{Baseline Details}

Unless otherwise specified, we run all baselines using their publicly released code and default configurations.

\noindent
\textbf{DA3.}
For Depth Anything 3 (DA3), we use its strongest released model, GIANT-LARGE1.1, for all evaluations.

\noindent
\textbf{COLMAP.}
For the T\&T PSNR evaluation, we use the COLMAP poses precomputed by ACE0~\citep{brachmann2024acezero} while for ETH3D evalution, we use the poses from AMB3R~\cite{wang2025amb3r} and MASt3R-SfM~\cite{murai2025mast3r}.

\noindent
\textbf{GLOMAP.}
We use the default configuration of GLOMAP~\cite{pan2024glomap}. For a fair comparison, the image matching stage uses the same image pairs as our pipeline. In a few cases, GLOMAP produces outlier poses with extremely large translations; for these frames, we replace the pose with the last valid estimate.

\noindent
\textbf{SAIL-Recon.}
Following the original SAIL-Recon protocol~\citep{dengli2025sail}, we refine the estimated poses with photometric bundle adjustment before evaluating PSNR on T\&T.

\noindent
\textbf{LingBot-Map.}
For LingBot-Map~\citep{chen2026geometric}, we use the author-recommended \texttt{lingbot-map-long} checkpoint with a keyframe interval of 1 by default. On the Courthouse scene, the default setting fails, so we rerun it with a keyframe interval of 2. For TUM RGB-D, since sequence lengths vary substantially, we evaluate keyframe intervals of 1, 2, and 3, and report the best result for each sequence.

\noindent
\textbf{LoGeR.}
For LoGeR~\citep{zhang2026loger}, we use LoGeR$^{*}$, the variant that adds a purely feed-forward alignment step to align raw predictions into a consistent global coordinate system. This setting also includes the chunk-alignment procedure used for long-sequence reconstruction.

\noindent
\textbf{VGGT-SLAM 2.0.}
For VGGT-SLAM 2.0~\citep{maggio2025vggt-slam2}, we do not skip frames during evaluation; all frames are kept and participate in the subsequent optimization.

\subsection{Additional Results}

\noindent
\textbf{Additional T\&T qualitative results.}
Fig.~\ref{fig:tntpsnr2} provides additional novel-view synthesis comparisons on Tanks and Temples. Our rendered images are consistently closer to the ground-truth views, preserving sharper object boundaries and more stable scene structures. In contrast, COLMAP~\cite{schonberger2016structure}, DA3~\cite{depthanything3}, and LingBot-Map~\cite{chen2026geometric} often produce visible artifacts, blur, or distorted geometry. These qualitative results further support the PSNR comparison in Table~\ref{tab:tnt_psnr_full}, showing that our estimated poses lead to more accurate neural rendering.

\noindent
\textbf{Additional KITTI trajectory visualizations.}
Fig.~\ref{fig:kitti03040506} shows trajectory visualizations on additional KITTI~\cite{Geiger2012kitti} sequences. Our method remains close to the ground-truth trajectories across both short and long driving sequences, while competing methods exhibit larger drift or unstable trajectory shapes. This further validates the robustness of our sliding-window association and track propagation strategy under forward-driving motion, where limited parallax makes long-range pose estimation challenging.

\noindent
\textbf{Dense reconstruction results.}
Fig.~\ref{fig:densepoints} visualizes dense point clouds recovered by our method on both ETH3D~\cite{schoeps2017eth3d} and Tanks and Temples~\cite{knapitsch2017tanks} scenes. The reconstructions preserve coherent scene layouts across indoor, outdoor, and object-centric environments, demonstrating that the optimized poses and recovered dense depths can be fused into globally aligned dense geometry. These results also show that our framework is not limited to sparse pose estimation, but can provide dense reconstruction suitable for downstream 3D applications.

\noindent
\textbf{Full T\&T quantitative results.}
Table~\ref{tab:tnt_psnr_full} reports the full PSNR results on all 19 Tanks and Temples scenes. Our method achieves the best average PSNR and obtains the highest score on most scenes. Compared with feed-forward models and streaming reconstruction methods, the consistent improvement across diverse scenes indicates that explicit multi-view tracks and global geometric refinement are important for high-fidelity pose estimation and rendering quality.

\noindent
\textbf{Runtime Analyse}
\input{tables/fps}
Table~\ref{tab:fps_comparison} compares the processing speed of different methods on an 800-frame sequence. Our method achieves 2.06 FPS, which remains practical despite including dense matching, sliding-window association, and global optimization. This indicates that the additional optimization stage does not make the pipeline prohibitively expensive. In contrast, traditional SfM pipelines such as COLMAP~\cite{schonberger2016structure} and GLOMAP~\cite{pan2024glomap} are significantly slower, mainly due to feature matching, geometric verification, and CPU-based optimization. Although some feed-forward or streaming methods achieve higher FPS, they generally sacrifice pose accuracy or global consistency. Overall, our method provides a favorable balance between efficiency and reconstruction accuracy.

\subsection{Failure Case Analysis.}
\begin{figure}
    \centering
    \includegraphics[width=1.00\linewidth]{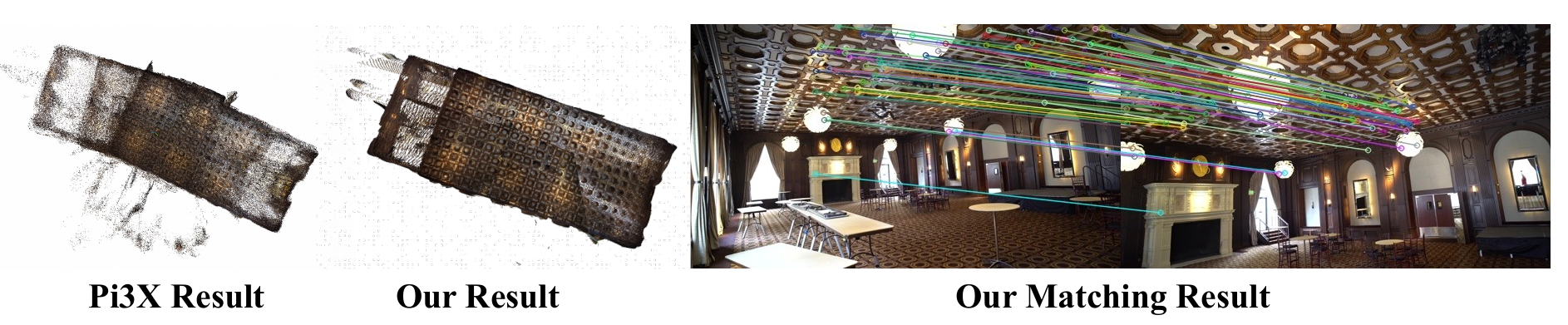}
    \caption{Failure case on the T\&T Ballroom~\cite{knapitsch2017tanks} scene. Ambiguous foundation-model geometry leads to incorrect matching and degraded reconstruction.}
    \label{fig:failure_ballroom}
    \vspace{-2mm}
\end{figure}

Fig.~\ref{fig:failure_ballroom} shows a typical failure case on the T\&T Ballroom~\cite{knapitsch2017tanks} scene. The initial Pi3X~\cite{wang2025pi3} prediction already contains strong geometric ambiguity: although the input images observe a single room, the predicted structure is separated into several inconsistent room-like fragments. Since our matching head is built upon the tokens produced by the Pi3X backbone, such ambiguous geometry can also affect the predicted correspondences. As shown in the matching visualization, visually repetitive structures, such as ceiling lamps, are incorrectly matched to different locations. Although our optimization can partially reduce the inconsistency, it cannot fully recover from severely incorrect initial geometry and corrupted matches. Similar failures are also observed in a few other T\&T scenes, such as Palace. This limitation suggests that future work should reduce the dependence of dense matching on imperfect foundation-model geometry and make the correspondence estimation more robust under ambiguous scene layouts.

%% file: tables/tnt_full.tex
\begin{table*}[t]
\centering
\caption{Full PSNR comparison on all 19 Tanks and Temples scenes. Best results are in bold.}
\label{tab:tnt_psnr_full}
\resizebox{\textwidth}{!}{
\begin{tabular}{lcccccccccccccccccccc}
\toprule
Scene 
& Auditorium & Ballroom & Barn & Caterpillar & Church & Courthouse & Courtroom & Family & Francis & Horse 
& Ignatius & Lighthouse & Meetingroom & Museum & Palace & Playground & Temple & Train & Truck & Avg. \\
\midrule
DA3 \cite{depthanything3}
& 19.40 & 13.41 & 21.78 & 15.73 & 16.51 & OOM & 16.34 & 18.93 & 20.38 & 18.60 
& 18.48 & 17.43 & 17.78 & 15.60 & 12.86 & 19.01 & 17.21 & 15.88 & 18.17 & 17.42 \\
COLMAP \citep{schonberger2016structure} 
& 19.60 & 16.34 & 24.09 & 17.04 & 18.14 & 18.04 & 18.25 & 19.40 & 21.80 & 19.47 
& 20.07 & 16.65 & 18.59 & 16.87 & 13.56 & 19.07 & 18.10 & \textbf{16.70} & 21.08 & 18.57 \\
GLOMAP \citep{pan2024glomap} 
& 11.67 & \textbf{20.52} & 24.26 & 16.97 & \textbf{18.20} & \textbf{21.08} & 18.20 & 13.08 & 22.00 & 14.85 
& 20.12 & 17.76 & 19.56 & 11.81 & 13.35 & 15.72 & 18.23 & 12.35 & 21.12 & 17.41 \\
SAILRecon \citep{dengli2025sail} 
& \textbf{20.30} & 14.80 & 23.50 & 16.80 & 17.00 & 15.09 & 17.40 & \textbf{20.60} & 21.80 & \textbf{20.10} 
& 19.50 & 18.20 & 19.50 & 15.40 & 14.30 & 20.30 & 17.80 & 16.20 & 20.90 & 18.39 \\
Scal3R \citep{xie2026scal3rscalabletesttimetraining}
& 16.55 & 13.56 & 20.00 & 14.86 & 15.08 & 18.20 & 13.28 & 17.21 & 16.08 & 16.58 
& 14.82 & 12.09 & 14.74 & 13.01 & 13.14 & 16.79 & 17.09 & 13.04 & 14.70 & 15.31 \\
Pi3X \citep{wang2025pi3} 
& 19.05 & 13.46 & 21.22 & 16.04 & 17.01 & 16.74 & 17.23 & 18.89 & 20.27 & 18.19 
& 18.16 & 17.34 & 17.63 & 15.56 & 12.47 & 18.20 & 16.83 & 15.57 & 18.67 & 17.29 \\
LingBot-Map \cite{chen2026geometric}
& 18.30 & 12.90 & 18.54 & 14.51 & 15.44 & 14.43 & 15.65 & 17.60 & 19.01 & 17.17 
& 16.59 & 15.29 & 16.08 & 14.62 & 12.23 & 17.18 & 15.28 & 14.77 & 16.01 & 15.87 \\
AMB3R \cite{wang2025amb3r} 
& 18.05 & 12.68 & 20.08 & 15.45 & 15.98 & 15.64 & 15.58 & 19.50 & 19.19 & 19.08 
& 18.07 & 15.00 & 16.63 & 15.97 & 11.77 & 18.04 & 15.74 & 15.66 & 18.97 & 16.69 \\
Ours 
& 20.13 & 17.81 & \textbf{24.50} & \textbf{17.25} & 18.01 & 20.72 & \textbf{18.31} & 20.27 
& \textbf{22.01} & 19.79 & \textbf{20.43} & \textbf{18.76} & \textbf{19.68} & \textbf{17.43} 
& \textbf{14.42} & \textbf{21.68} & \textbf{18.82} & 16.60 & \textbf{21.13} & \textbf{19.36} \\
\bottomrule
\end{tabular}
}
\end{table*}

%% file: tables/fps.tex
\begin{table*}[h]
\centering
\caption{Comparison of processing speed across different methods on an 800-frame sequence. We report FPS, where higher values indicate faster performance.}
\label{tab:fps_comparison}
\resizebox{\textwidth}{!}{
\begin{tabular}{lcccccccccccc}
\toprule
\multirow{2}{*}{Metric} 
& \multicolumn{2}{c}{Traditional SfM}
& \multicolumn{3}{c}{Feed-forward}
& \multicolumn{5}{c}{Streaming / Chunk + Optimization}
& \multicolumn{1}{c}{Ours} \\
\cmidrule(lr){2-3}
\cmidrule(lr){4-6}
\cmidrule(lr){7-11}
\cmidrule(lr){12-12}
& COLMAP~\cite{schonberger2016structure}
& GLOMAP~\cite{pan2024glomap}
& Pi3X~\cite{wang2025pi3}
& DA3~\cite{depthanything3}
& FastVGGT~\cite{shen2025fastvggt}
& VGGT-Long~\cite{deng2025vggtlongchunkitloop}
& VGGT-SLAM~\cite{maggio2025vggt}
& MASt3R-SLAM~\cite{murai2025mast3r}
& LingBot-Map~\cite{chen2026geometric}
& Scal3R~\cite{xie2026scal3rscalabletesttimetraining}
& Ours \\
\midrule
FPS $\uparrow$
& 0.14
& 0.67
& 1.64
& 8.10
& 15.10
& 4.00
& 16.87
& 6.06
& 6.80
& 2.20
& 2.06 \\
\bottomrule
\end{tabular}
}
\end{table*}